\documentclass[10pt,twocolumn,letterpaper]{article}

\usepackage[pagenumbers]{cvpr} 

\usepackage{pifont}    
\usepackage{makecell} 
\usepackage{multirow} 
\usepackage[accsupp]{axessibility}  

\usepackage{algorithm}
\usepackage{algorithmic}
\usepackage{listings}
\newcommand{\TODO}[1]{\textbf{\color{red}[TODO: #1]}}
\renewcommand{\TODO}[1]{}

\definecolor{cvprblue}{rgb}{0.21,0.49,0.74}
\usepackage[pagebackref,breaklinks,colorlinks,allcolors=cvprblue]{hyperref}

\def\confName{CVPR}
\def\confYear{2026}

\title{Guiding Diffusion Models with Semantically Degraded Conditions}

\author{
Shilong Han$^*$\quad Yuming Zhang$^*$\quad Hongxia Wang$^\dagger$ \\
College of Science, National University of Defense Technology \\
{\tt\small hanshilong20@nudt.edu.cn, zhangyuming@nudt.edu.cn, wanghongxia@nudt.edu.cn}
}

\begin{document}
\maketitle
{\renewcommand\thefootnote{}\footnotetext{\parbox{\linewidth}{$^*$ These authors contributed equally.\\$^\dagger$ Corresponding author.}}}
\begin{abstract}
    Classifier-Free Guidance (CFG) is a cornerstone of modern text-to-image models, yet its reliance on a semantically vacuous null prompt ($\varnothing$) generates a guidance signal prone to geometric entanglement. This is a key factor limiting its precision, leading to well-documented failures in complex compositional tasks. We propose Condition-Degradation Guidance (CDG), a novel paradigm that replaces the null prompt with a strategically degraded condition, $\boldsymbol{c}_{\text{deg}}$. This reframes guidance from a coarse ``good vs. null" contrast to a more refined ``good vs. almost good" discrimination, thereby compelling the model to capture fine-grained semantic distinctions. We find that tokens in transformer text encoders split into two functional roles: content tokens encoding object semantics, and context-aggregating tokens capturing global context. By selectively degrading only the former, CDG constructs $\boldsymbol{c}_{\text{deg}}$ without external models or training. Validated across diverse architectures including Stable Diffusion 3, FLUX, and Qwen-Image, CDG markedly improves compositional accuracy and text-image alignment. As a lightweight, plug-and-play module, it achieves this with negligible computational overhead. Our work challenges the reliance on static, information-sparse negative samples and establishes a new principle for diffusion guidance: the construction of adaptive, semantically-aware negative samples is critical to achieving precise semantic control. Code is available at \url{https://github.com/Ming-321/Classifier-Degradation-Guidance}.
\end{abstract}    
\section{Introduction}
\label{sec:intro}

\begin{figure*}[t]
  \centering
  \includegraphics[width=\textwidth]{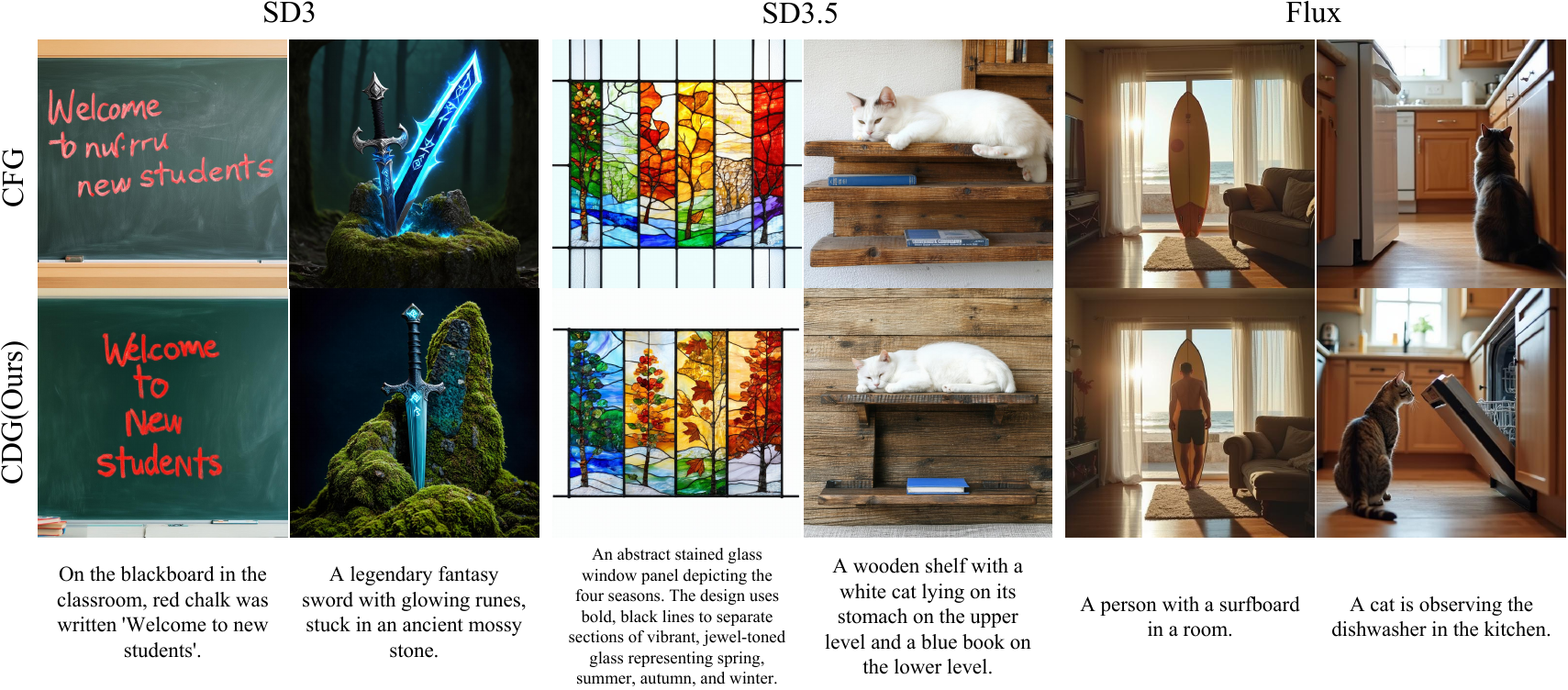}
  \caption{\textbf{Qualitative comparison between Classifier-Free Guidance (CFG) and our Condition-Degradation Guidance (CDG) across three state-of-the-art models (SD3, SD3.5, and Flux).} These examples demonstrate CDG's superior capability in handling complex compositional prompts where CFG often fails. CDG consistently outperforms CFG in accurate text rendering, precise spatial relationships and attribute binding, as well as complex object interactions.}
  \label{fig:title_image}
\end{figure*}

Diffusion Models (DMs) have become a dominant force in generative modeling~\cite{ho2020denoising, song2019generative, song2021scorebased}, with latent diffusion~\cite{rombach2022high} and transformers~\cite{peebles2023scalable} continually advancing text-to-image synthesis. Pivotal to this progress is Classifier-Free Guidance (CFG)~\cite{ho2022classifier}, which steers generation by extrapolating unconditional predictions toward conditional ones, and has become a cornerstone of modern text-to-image systems~\cite{hertz2022prompt, couairon2022diffedit, zhang2023adding, brooks2023instructpix2pix, dai2023emu, sheynin2023emuedit, kawar2023imagic, huang2023composer, zhao2023uni, mou2024t2i}.

While pivotal, CFG exhibits failure modes in compositional tasks—text rendering, complex attribute binding, and spatial relationships (\cref{fig:title_image}). We argue this stems from the semantic poverty of $\emptyset$: the large gap between $\boldsymbol{c}$ and $\emptyset$ yields an entangled guidance signal that mixes content generation with style and structure~\cite{sadat2025eliminating,tcfg2025}. In contrast, a semantically close $\boldsymbol{c}_{\text{deg}}$ enables common-mode rejection—suppressing shared components to isolate pure semantic corrections, as validated in \cref{sec:mechanism_analysis}.

Existing methods to address CFG's limitations fall into two camps. \emph{Process Rectification} methods~\cite{sadat2025eliminating,tcfg2025,song2025rethinking,chen2025guidance} retain $\boldsymbol{c}$ vs.\ $\emptyset$ but apply post-hoc corrections—treating symptoms, not causes. Meanwhile, \emph{Negative Reframing} methods~\cite{karras2024guiding,bai2025weak,tpg2025,ahn2024selfrectifying,sadat2024no,sadat2024cads,azarian2024sfg,desai2024dnp} explore alternatives to $\emptyset$—using weak models, random perturbations, attention manipulations, or VLM-generated negatives—yet none exploit the inherent semantic structure within the prompt's own token embeddings. A critical question remains: \emph{how to construct a negative condition that adaptively degrades the core semantics of the positive prompt in a principled manner?}

We address this challenge through a key structural observation: in transformer-based text encoders, token embeddings naturally divide into \emph{content tokens} (encoding object-specific semantics) and \emph{context-aggregating tokens} (encoding global compositional context). By selectively removing content tokens while preserving context-aggregating tokens, a strategy we call \emph{stratified degradation}, we construct a degraded condition $\boldsymbol{c}_{\text{deg}}$ that retains the prompt's global semantic scaffold while losing fine-grained details, reframing guidance from ``good vs.\ null'' to ``good vs.\ \emph{almost good}''. We instantiate this principle in Condition-Degradation Guidance (CDG), a lightweight, plug-and-play module (\cref{fig:title_image}).

Stratified degradation rests on the role of context-aggregating tokens: padding and special tokens that originally lack intrinsic semantics but acquire rich global context through attention. This is not a quirk of specific architectures but a fundamental property of transformer encoders; we confirm generality across diverse architectures (\cref{ssec:sota}).

We validate CDG on Stable Diffusion 3, SD3.5, FLUX.1-dev, and Qwen-Image, demonstrating consistent improvements over baselines on FID, CLIP Score, VQA Score, and GenAI-Bench compositional reasoning with minimal overhead.

In summary, our contributions are:
\begin{itemize}
    \item We reveal a functional dichotomy in transformer text encoders between \emph{content tokens} and \emph{context-aggregating tokens}, and propose \emph{stratified degradation} as a principled strategy for constructing semantically degraded negative conditions.
    \item Based on this finding, we introduce Condition-Degradation Guidance (CDG), a lightweight, training-free, plug-and-play module requiring no external models or additional training.
    \item Extensive experiments across diverse models (SD3, SD3.5, FLUX.1-dev, Qwen-Image) validate CDG, providing geometric evidence for superior signal orthogonality and demonstrating consistent metric improvements with negligible overhead.
\end{itemize}

\section{Related Work}
\label{sec:related_work}

Our work is situated within a broad research effort to enhance text-to-image generation. We contextualize CFG refinements~\citep{ho2022classifier}, then focus on paradigms moving beyond the null prompt.

\textbf{Refinement of the CFG Framework.} Many methods focus on refining how the standard $\boldsymbol{c}$ vs $\emptyset$ guidance is applied, via geometric corrections (APG~\citep{sadat2025eliminating}) or SVD (TCFG~\citep{tcfg2025}). While impactful, this research retains the foundational reliance on the semantically poor null condition.

\textbf{Beyond the Null Prompt.} Another paradigm moves beyond $\emptyset$ entirely, reframing guidance as a contrast between a ``good'' prediction and a ``degraded'' one. These methods can be distinguished by the source of degradation:
\begin{itemize}
    \item \textbf{Model Level.} One approach uses a separate, typically weaker model to provide the negative signal, such as in Autoguidance~\citep{karras2024guiding} and Weak-to-Strong Diffusion~\citep{bai2025weak}—effective but requires external model tuning.
    
    \item \textbf{Internal Mechanism Level.} Another direction perturbs the model's internal representations during the forward pass, including perturbing attention matrices (PAG~\citep{ahn2024selfrectifying}) or smoothing energy curvature (SEG~\citep{hong2024smoothed}). These methods manipulate the model's computational flow to generate an implicit negative signal. As they operate on a different principle from input-level modifications, they are largely orthogonal to our approach and can potentially be combined.
    
    \item \textbf{Input Level.} The most direct strategy is to degrade the conditioning signal $\boldsymbol{c}$ itself. Methods in this category include using random prompts (ICG~\citep{sadat2024no}), adding unstructured Gaussian noise (CADS~\citep{sadat2024cads}), spatially varying negatives (SFG~\citep{azarian2024sfg}), and VLM-generated negatives (DNP~\citep{desai2024dnp}). However, these approaches either remain semantically blind or require expensive external models, without exploiting the inherent semantic structure within the prompt's own token embeddings.
\end{itemize}

\textbf{Our Approach.} To address this gap, our Condition-Degradation Guidance (CDG) exploits the inherent functional dichotomy in transformer text encoders between content tokens and context-aggregating tokens. Through \emph{stratified degradation}---selectively removing content tokens while preserving context-aggregating tokens---CDG constructs a semantically degraded condition that retains global compositional context. This creates a precise ``good'' vs.\ ``almost good'' contrast, directly improving guidance quality without external models or blind perturbations.

\section{Background} \label{sec:preliminaries}

\textbf{Denoising Diffusion.} Denoising diffusion generates samples from a data distribution $p_{\text{data}}$ by reversing a process that gradually corrupts the data with Gaussian noise. 
This process yields smoothed densities $p(\boldsymbol{x}; \sigma) = p_{\text{data}}(\boldsymbol{x}) * \mathcal{N}(\mathbf{0}, \sigma^2\mathbf{I})$, indexed by the noise level $\sigma$. 
The generation process is then defined by a probability flow ODE \cite{karras2022edm, song2021denoising, song2021scorebased} that evolves samples from pure noise back towards the data distribution: 
\begin{equation}
    d\boldsymbol{x}_\sigma = -\sigma \nabla_{\boldsymbol{x}_\sigma} \log p(\boldsymbol{x}_\sigma; \sigma) d\sigma.
\end{equation}
The core of this process is the score function, $\nabla_{\boldsymbol{x}_\sigma} \log p(\boldsymbol{x}_\sigma; \sigma)$, which directs the update. 

The score is approximated by a neural network $D_\theta(\boldsymbol{x}_\sigma; \sigma)$ parameterized by weights $\theta$. 
While this network can be parameterized in various ways (e.g., to predict noise), we follow the formulation where it is trained as a denoiser to predict the clean sample $\boldsymbol{x}_0$ from a noised input $\boldsymbol{x}_\sigma$: 
\begin{equation}
    \theta^* = \arg\min_{\theta} \mathbb{E}_{\boldsymbol{x}_0, {\sigma}, \boldsymbol{\epsilon}} \left[ \left\| D_\theta(\boldsymbol{x}_0 + \sigma\boldsymbol{\epsilon}; \sigma) - \boldsymbol{x}_0 \right\|^2 \right].
\end{equation}
The expectation is taken over $\boldsymbol{x}_0 \sim p_{\text{data}}$, $\boldsymbol{\epsilon} \sim \mathcal{N}(\mathbf{0,I})$, and $\sigma \sim p_{\text{train}}$, where $p_{\text{train}}$ governs the noise level distribution during training. 
Given $D_\theta$, we can estimate of the score function $\nabla_{\boldsymbol{x}_\sigma} \log p(\boldsymbol{x}_\sigma; \sigma)\approx(D_\theta(\boldsymbol{x}_\sigma;\sigma)-\boldsymbol{x}_\sigma)/\sigma^2$. 
For conditional generation, the network is trained with an additional conditioning input $\boldsymbol{c}$ (e.g., a text embedding), becoming $D_\theta(\boldsymbol{x}_\sigma; \sigma, \boldsymbol{c})$, which provides an estimate of the conditional score $\nabla_{\boldsymbol{x}_\sigma} \log p(\boldsymbol{x}_\sigma| \boldsymbol{c}; \sigma)$. 

\textbf{Classifier-Free Guidance.} Due to approximation errors 
inherent in finite-capacity networks, generated images often fail to match the fidelity of the training data. 

A widely adopted technique to counteract this is Classifier-Free Guidance (CFG) \cite{ho2022classifier}, which enhances sample quality by extrapolating from an unconditional prediction towards a conditional one: 
\begin{align}
D_\theta^{\text{CFG}}(\boldsymbol{x}_\sigma; \sigma, \boldsymbol{c}) &= D_\theta(\boldsymbol{x}_\sigma; \sigma, \boldsymbol{c}) \label{eq:cfg} \\
&+ (w-1) (D_\theta(\boldsymbol{x}_\sigma; \sigma, \boldsymbol{c}) - D_\theta(\boldsymbol{x}_\sigma; \sigma, \emptyset)), \notag
\end{align}
where $w > 1$ is the guidance scale. 

Recalling the equivalence between the denoiser and the score function, we can rewrite the above equation as:
\begin{align}
    \nabla_{\boldsymbol{x}_\sigma}\log p_w(\boldsymbol{x}_\sigma|\boldsymbol{c};\sigma)
    &= \nabla_{\boldsymbol{x}_\sigma}\log p(\boldsymbol{x}_\sigma|\boldsymbol{c};\sigma)  \\
    &\quad + (w-1)\nabla_{\boldsymbol{x}_\sigma}\log\frac{p(\boldsymbol{x}_\sigma|\boldsymbol{c};\sigma)}{p(\boldsymbol{x}_\sigma|\emptyset;\sigma)}. \notag
\end{align}
\section{Understanding CDG: A Geometric Perspective}
\label{sec:mechanism_analysis}

We propose Condition-Degradation Guidance (CDG), which replaces the semantically distant null condition $\emptyset$ in CFG with a semantically degraded condition $\boldsymbol{c}_{\text{deg}}$ that is close to the original prompt $\boldsymbol{c}$:
\begin{align}
\label{eq:cdg_formulation}
D_\theta^{\text{CDG}}(\boldsymbol{x}_\sigma; \sigma, \boldsymbol{c}) &= D_\theta(\boldsymbol{x}_\sigma; \sigma, \boldsymbol{c}) \\
&+ (w-1) (D_\theta(\boldsymbol{x}_\sigma; \sigma, \boldsymbol{c}) - D_\theta(\boldsymbol{x}_\sigma; \sigma, \boldsymbol{c}_{\text{deg}})). \notag
\end{align}
This reframes guidance from a coarse ``good vs.\ null'' contrast to a refined ``good vs.\ almost good'' discrimination. But why does this substitution improve guidance quality? 
We hypothesize that semantically distant contrasts ($\boldsymbol{c}$ vs. $\emptyset$) may produce guidance signals that interfere with the primary denoising direction, while semantic differencing ($\boldsymbol{c}$ vs. $\boldsymbol{c}_{\text{deg}}$) achieves better decoupling. To test this hypothesis, we analyze the geometric properties of the guidance signals.

\textbf{Analytical Framework.} Our analysis builds on the manifold hypothesis~\cite{Bengio2013,Fefferman2016}, which posits that high-dimensional natural image data resides on low-dimensional manifolds. The diffusion process can be viewed as an evolution across a series of progressively smoother manifolds $\mathcal{M}_t$. Recent theoretical and empirical work~\cite{Stanczuk2024,tcfg2025} has demonstrated that the score function $\nabla\log p_t(z_t)$ aligns with the manifold's normal space $\mathcal{N}_{z_t}(\mathcal{M}_t)$, which governs the primary denoising direction, throughout the reverse process. Following~\cite{tcfg2025}, we apply SVD to conditional predictions $\{\boldsymbol{\varepsilon}_{\boldsymbol{c}}\}$ across diverse prompts from MS-COCO 2017~\cite{lin2014microsoft} to approximate the principal denoising subspace $\mathcal{S}_{\boldsymbol{c}}(t)$ at each timestep $t$.

To quantify the geometric relationship between guidance signals and $\mathcal{S}_{\boldsymbol{c}}(t)$, we introduce two metrics, where $\mathcal{S}_g$ denotes the subspace spanned by the guidance signal $\Delta\boldsymbol{\varepsilon}$:
\begin{itemize}
    \item \textbf{Geometric Decoupling} measures orthogonality between $\mathcal{S}_g$ and $\mathcal{S}_{\boldsymbol{c}}$:
    \begin{equation}
    \text{Decoupling}(\mathcal{S}_g, \mathcal{S}_{\boldsymbol{c}}) = \frac{1}{k} \sum_{i=1}^k \sin^2(\theta_i),
    \end{equation}
    where $\theta_i$ is the $i$-th principal angle between the two subspaces and $k$ is the subspace dimension; values approaching 1 indicate near-perfect orthogonality.
    
    \item \textbf{Interference Energy Ratio} measures the fraction of guidance energy projected onto $\mathcal{S}_{\boldsymbol{c}}(t)$:
    \begin{equation}
    \text{Interference}(\Delta\boldsymbol{\varepsilon}) = \frac{\| P_{\mathcal{S}_{\boldsymbol{c}}(t)} \Delta\boldsymbol{\varepsilon} \|_F^2}{\| \Delta\boldsymbol{\varepsilon} \|_F^2},
    \end{equation}
    where $P_{\mathcal{S}_{\boldsymbol{c}}(t)}$ denotes the orthogonal projection onto $\mathcal{S}_{\boldsymbol{c}}(t)$; lower values indicate less interference with denoising.
\end{itemize}

\begin{figure}[t]
    \centering
    \includegraphics[width=0.47\textwidth]{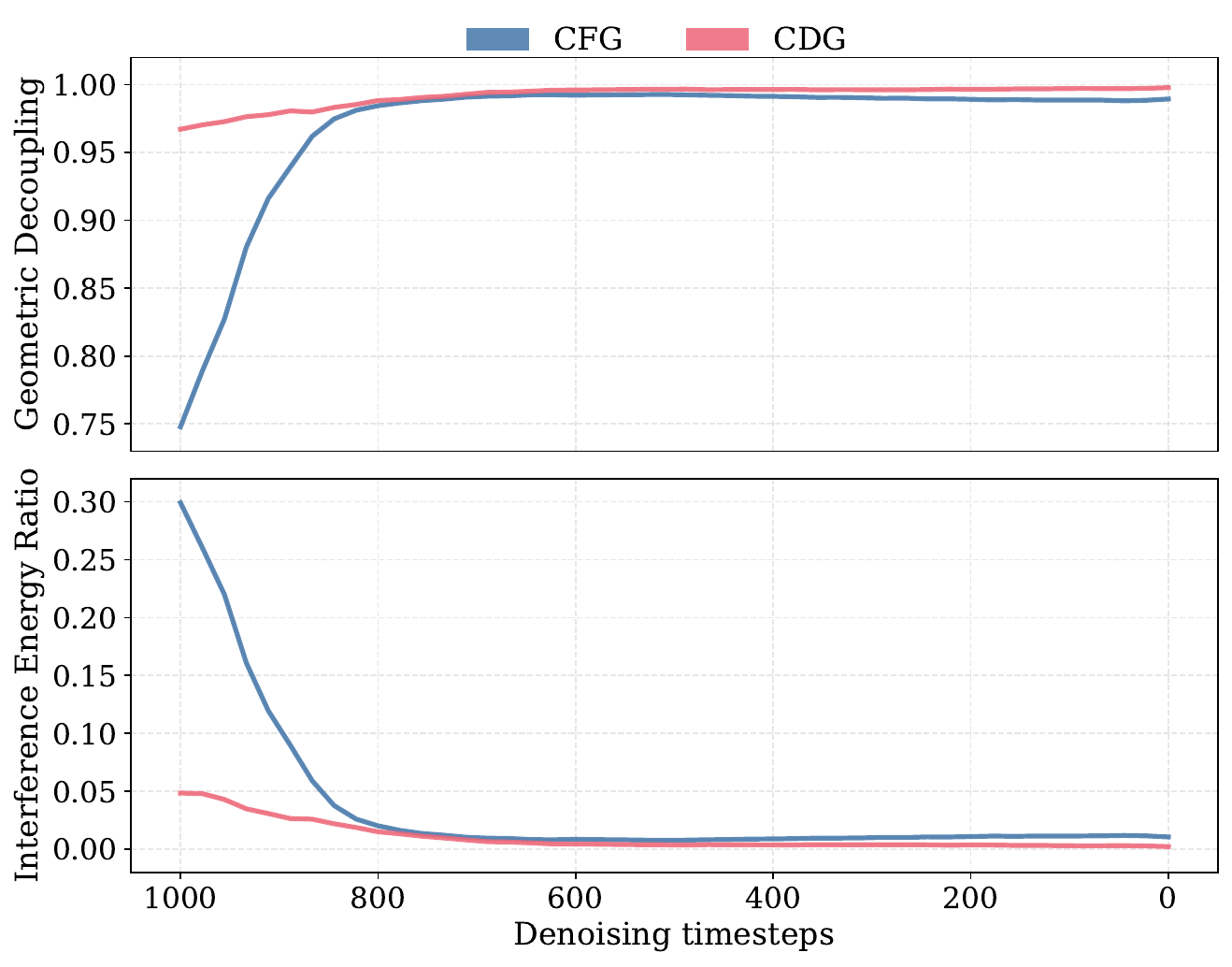}
    \caption{
        \textbf{CDG synthesizes a geometrically superior guidance signal compared to CFG.} 
    (Top) Geometric Decoupling: CDG maintains near-perfect orthogonality throughout generation, while CFG suffers from significant early-stage entanglement. 
    (Bottom) Interference Energy Ratio: CDG exhibits minimal interference, in stark contrast to CFG's substantial energy waste in misaligned directions. 
    Together, these analyses demonstrate that CDG's guidance signal is structurally cleaner and more efficient from its inception, explaining its enhanced compositional control.
    }
    \label{fig:four_plots}
\end{figure}

We evaluate these metrics on CFG and CDG across multiple diffusion timesteps, and the results in \cref{fig:four_plots} reveal a striking divergence.

As \cref{fig:four_plots} shows, CDG maintains near-perfect orthogonality with minimal interference throughout generation, directly supporting our hypothesis. This aligns with Sadat et al.~\cite{sadat2025eliminating}, who observed that perpendicular guidance components improve quality while parallel ones introduce artifacts.

\textbf{Why does CDG achieve this?} We attribute it to a common-mode rejection effect. As semantic neighbors, $\boldsymbol{c}$ and $\boldsymbol{c}_{\text{deg}}$ share similar normal components; their difference $\Delta\boldsymbol{\varepsilon}_{\text{CDG}} \propto \nabla_{z_t} \log \frac{p_t(z_t \mid \boldsymbol{c})}{p_t(z_t \mid \boldsymbol{c}_{\text{deg}})}$ cancels these, leaving primarily semantic distinctions. In contrast, CFG's semantically distant contrast ($\boldsymbol{c}$ vs. $\emptyset$) cannot achieve this cancellation, leading to entangled signals that conflate correction with denoising.

We hypothesize that the effectiveness of this ``common-mode rejection'' may stem from ensuring that $\boldsymbol{c}_{\text{deg}}$ preserves the ``global context'' shared with $\boldsymbol{c}$ (the common mode) while removing ``specific semantics'' (the correction signal). As we demonstrate in \cref{sec:method} (see \cref{fig:token_importance_visualization}), such a decoupling is achievable through our stratified degradation strategy. In \cref{sssec:hyperparams} and \cref{sssec:cfg_star}, we will observe asymmetric patterns in experimental results that are consistent with this geometric interpretation.

\section{Method}
\label{sec:method}

\begin{figure*}[t]
    \centering
    \includegraphics[width=1.0\textwidth]{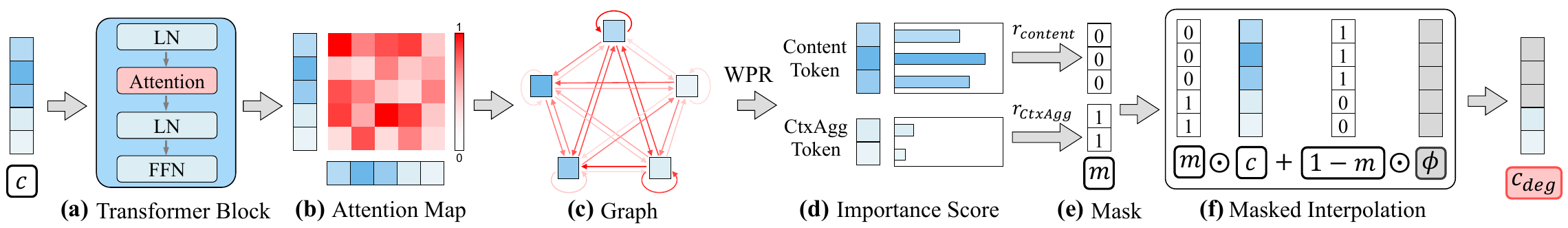}
    \caption{\textbf{An illustration of our proposed pipeline for constructing the semantically degraded condition, $\boldsymbol{c}_{\text{deg}}$.} The process begins with attention graph extraction (a--c), where the self-attention map (b) from a transformer block (a) is modeled as a graph (c). Next, the Weighted PageRank (WPR) algorithm is applied to compute an importance score for each token (d). Following our Stratified Degradation strategy, these scores are used to generate a binary mask $\boldsymbol{m}$ (e). Finally, the mask facilitates the construction of $\boldsymbol{c}_{\text{deg}}$ via masked interpolation (f) between the original condition $\boldsymbol{c}$ and the null condition $\emptyset$.}
    \label{fig:pipeline}
    \end{figure*}

To construct $\boldsymbol{c}_{\text{deg}}$ as introduced in \cref{sec:mechanism_analysis}, we analyze the model's internal information flow to identify and strategically degrade the most semantically important tokens in the original prompt $\boldsymbol{c}$. Our complete pipeline is illustrated in \cref{fig:pipeline}.

\begin{figure*}[t]
    \centering
    \includegraphics[width=1.0\linewidth]{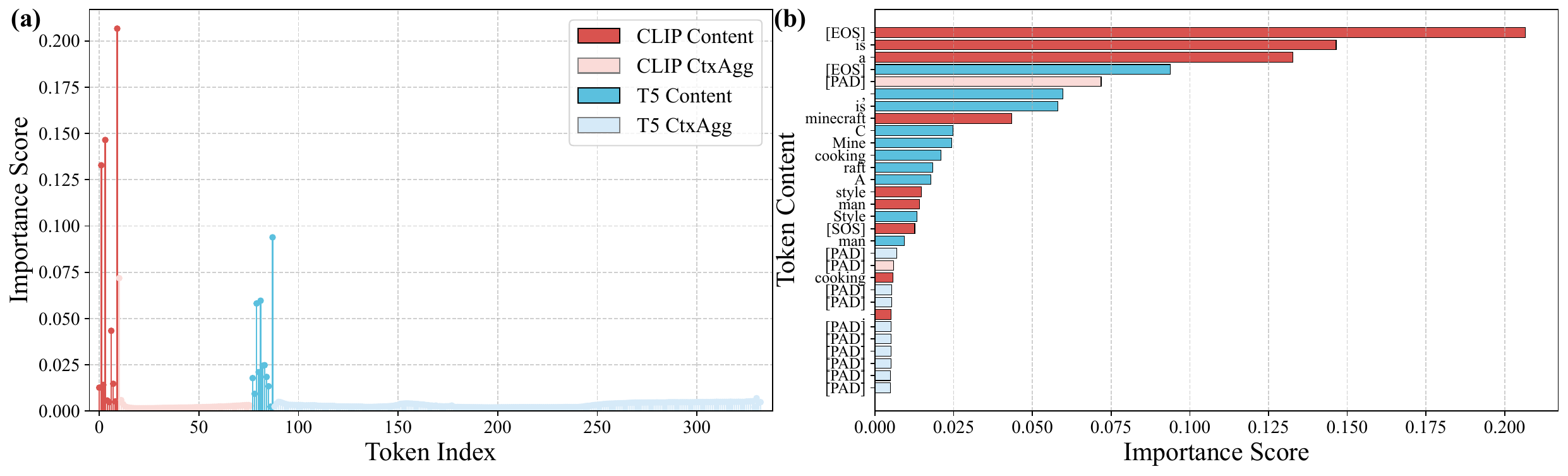}
    \caption{\textbf{WPR reveals a clear importance dichotomy between content and context-aggregating tokens:} content tokens carry fine-grained semantics while context-aggregating tokens carry coarse-grained semantics, as exemplified by the prompt ``A man is cooking, MineCraft Style.'' (a) The stem plot shows that high importance scores (red for CLIP, cyan for T5) are almost exclusively concentrated on semantic content tokens. (b) The ranked list confirms that the top tokens (``minecraft'', ``cooking'', ``man'') are almost all content-related. This dichotomy motivates our Stratified Degradation strategy, which first degrades content tokens and then context-aggregating tokens for controllable semantic degradation.}
    \label{fig:token_importance_visualization}
\end{figure*}

To validate the content/context-aggregating dichotomy that motivates stratified degradation, we employ Weighted PageRank (WPR) as an analytical tool for quantifying token importance. Unlike standard approaches that aggregate cross-attention scores—which can paradoxically assign higher importance to context-aggregating tokens (see Appendix~A)—we model the token relationships as a graph, as shown in \cref{fig:pipeline} (a--c). Specifically, at a designated block $\lambda_{\text{block}}$ (detailed below), we extract the self-attention map $A \in \mathbb{R}^{N \times N}$, where $N$ is the sequence length. The tokens serve as the graph's nodes, while the attention weights in $A$ provide the edge weights. We then apply the Weighted PageRank (WPR) algorithm~\cite{xing2004weighted,wang2024zero} to this attention-weighted graph to compute a final importance score vector $\boldsymbol{s} \in \mathbb{R}^{N}$ (\cref{fig:pipeline} (d)). The core iterative update of WPR is defined as: 
\begin{equation}
    \boldsymbol{s}^{(k+1)} = \frac{A^T \boldsymbol{s}^{(k)}}{\|A^T \boldsymbol{s}^{(k)}\|_1},
    \label{eq:wpr}
\end{equation}
where $\boldsymbol{s}^{(k)}$ is the importance score vector at the $k$-th iteration, and the process is repeated until convergence. Details are provided in Appendix~A.

As shown in \cref{fig:token_importance_visualization}, WPR reveals a clear importance dichotomy: content tokens exhibit substantially higher importance scores than context-aggregating tokens. This separation is intuitive: content tokens (e.g., ``minecraft'', ``cooking'') from meaningful text carry specific, fine-grained semantics; context-aggregating tokens lack explicit content before encoding. Despite this, they absorb contextual information through the encoder, carrying coarse-grained global semantics (as verified in \cref{sssec:cfg_star}). The dichotomy revealed by WPR informs our design of the Stratified Degradation strategy: rather than degrading blindly, such as setting global degradation ratios, we exploit this structure to design a degradation path that prioritizes content tokens over context-aggregating tokens.

As shown in \cref{fig:pipeline}(e), we partition the set of token indices $\mathcal{T}$ into a content set $\mathcal{T}_{\text{content}}$ and a context-aggregating set $\mathcal{T}_{\text{CtxAgg}}$. We then introduce two key hyperparameters, $r_{\text{content}} \in [0, 1]$ and $r_{\text{CtxAgg}} \in [0, 1]$, which represent the desired replacement ratios for the content and context-aggregating tokens, respectively. We parameterize these two ratios through a single unified Degradation Ratio $R_{\text{deg}} \in [0, 2]$, which maps to the per-type ratios via:
\begin{equation}
    r_{\text{content}} = \min(R_{\text{deg}}, 1.0), \quad r_{\text{CtxAgg}} = \max(R_{\text{deg}} - 1.0, 0).
    \label{eq:unified_ratio}
\end{equation}
This formulation ensures that semantically important content tokens are degraded before context-aggregating tokens. Based on these ratios, the number of top-ranked tokens to replace from each subset is determined: $k_{\text{content}} = \lfloor r_{\text{content}} \cdot |\mathcal{T}_{\text{content}}| \rfloor$ and $k_{\text{CtxAgg}} = \lfloor r_{\text{CtxAgg}} \cdot |\mathcal{T}_{\text{CtxAgg}}| \rfloor$. The final binary replacement mask $\boldsymbol{m} \in \{0, 1\}^N$ is then defined for each token $i$ as:
\begin{equation}
m_i = \begin{cases} 
    0 & \text{if } i \in \mathcal{T}_{\text{content}} \text{ and } \text{rank}_i \le k_{\text{content}}, \\
    0 & \text{if } i \in \mathcal{T}_{\text{CtxAgg}} \text{ and } \text{rank}_i \le k_{\text{CtxAgg}}, \\
    1 & \text{otherwise},
\end{cases}
\label{eq:mask}
\end{equation}
where $\text{rank}_i$ is the rank of token $i$ within its respective subset, determined by sorting the corresponding importance scores $s_i$ in descending order. A smaller rank value (e.g., 1) thus signifies higher importance.

Through this design, $R_{\text{deg}}=1.0$ represents a natural ``semantic boundary'' that separates two degradation regimes:
\begin{itemize}
\item $R_{\text{deg}} \in [0, 1.0]$: removes content tokens (fine-grained semantics),
\item $R_{\text{deg}} \in (1.0, 2.0]$: removes context-aggregating tokens (coarse-grained semantics).
\end{itemize}
This dichotomy-driven formulation provides an interpretable control space. As we show in \cref{sssec:hyperparams} (\cref{fig:ablation_hyperparams}) and \cref{sssec:cfg_star} (\cref{fig:cfg_star_analysis}), $R_{\text{deg}}=1.0$ serves as a robust default across models—balancing multiple metrics while offering computational efficiency (no WPR computation needed at this boundary). Users can adjust around $R_{\text{deg}}=1.0$ to fine-tune style-alignment trade-offs.

Using the final mask $\boldsymbol{m}$, we construct the degraded condition $\boldsymbol{c}_{\text{deg}}$ by performing a masked interpolation between the original condition $\boldsymbol{c}$ and the null condition $\emptyset$, as shown in \cref{fig:pipeline} (f):
\begin{equation}
    \boldsymbol{c}_{\text{deg}} = \boldsymbol{m} \odot \boldsymbol{c} + (1 - \boldsymbol{m}) \odot \emptyset,
    \label{eq:c_deg}
\end{equation}
where $\odot$ denotes the element-wise Hadamard product. To make this degradation adaptive, we introduce an intervention block index, $\lambda_{\text{block}}$, to specify from which transformer block the attention map is extracted. At each step, reaching $\lambda_{\text{block}}$ triggers mask construction via \cref{eq:mask}, applying $\boldsymbol{c}_{\text{deg}}$ (\cref{eq:c_deg}) for all subsequent blocks within that step.

For computational efficiency, we compute the mask $\boldsymbol{m}$ only once at the first denoising step and reuse it throughout generation (\cref{tab:ablation_efficiency}), introducing minimal overhead with negligible performance impact.

\section{Experiments}
\label{sec:experiments}
\subsection{Experimental Setup}
\label{ssec:setup}

\textbf{Base Model.}
Our experiments are built upon four text-to-image diffusion models: Stable Diffusion 3 Medium (SD3)~\cite{esser2024scaling}, Stable Diffusion 3.5 (SD3.5)~\cite{esser2024scaling}, FLUX.1-dev (FLUX.1)~\cite{flux2024}, and Qwen-Image~\cite{wu2025qwenimage}. All models are based on Transformer architectures at their core.

\textbf{Dataset.}
We use 5,000 captions from the MS-COCO 2017 validation set~\cite{lin2014microsoft} for comprehensive assessment.

\textbf{Evaluation Metrics.}
To evaluate the performance of text-to-image models, we adopt four key metrics: FID~\cite{heusel2017gans}, Aesthetic Score~\cite{zhai2023sigmoid}, CLIP Score~\cite{radford2021learning}, and VQA Score~\cite{lin2024evaluating}. These metrics primarily assess performance from two aspects: image quality and text-image alignment.

Additionally, we report results on GenAI-Bench~\cite{li2024genai}, a compositional reasoning benchmark covering diverse basic and advanced skills, to evaluate complex compositional capabilities.

More details of models, datasets, hyperparameters, and metrics are provided in Appendix~C.1--C.3.

\subsection{Comparison with Baselines}
\label{ssec:sota}

\begin{table}[h]
    \centering
    \caption{Quantitative comparison on the MS-COCO 2017 validation set. Results are shown for SD3, SD3.5, FLUX.1, and Qwen-Image. Best results in \textbf{bold}, second-best \underline{underlined}. ($\downarrow$) indicates lower is better, ($\uparrow$) higher is better.}
    \small
    \begin{tabular}{cccccc}
    \toprule
    \multicolumn{2}{c}{\makecell{\textbf{Method}}} & \makecell{\textbf{FID}\\}$\downarrow$ & \makecell{\textbf{CLIP}\\\textbf{Score}}$\uparrow$ & \makecell{\textbf{Aesthetic}\\\textbf{Score} }$\uparrow$ & \makecell{\textbf{VQA}\\\textbf{Score} }$\uparrow$ \\
    \midrule
    \multirow{8}{*}{\textbf{SD3}} & CFG & 35.69 & \underline{31.73} & 5.66 & \underline{91.44} \\
     & CADS & 36.16 & 31.72 & 5.65 & \underline{91.44} \\
     & ICG & 39.09 & 31.41 & \textbf{5.79} & 90.89 \\
     & SEG & 41.90 & 30.28 & 5.56 & 84.15 \\
     & PAG & 50.60 & 30.15 & 5.52 & 81.27 \\
     & SFG & 38.92 & 31.72 & 5.52 & 90.52 \\
     & DNP & \underline{34.68} & 31.30 & 5.51 & 89.65 \\
     & \textbf{CDG} & \textbf{34.05} & \textbf{32.00} & \underline{5.70} & \textbf{92.40} \\
    \midrule
    \multirow{6}{*}{\textbf{SD3.5}} & CFG & \underline{34.56} & 31.85 & 6.21 & \underline{91.94} \\
     & CADS & 34.58 & \underline{31.86} & 6.21 & 91.83 \\
     & ICG & 35.41 & 31.70 & \textbf{6.32} & 91.80 \\
     & SEG & 38.90 & 30.71 & 6.16 & 88.49 \\
     & PAG & 39.70 & 30.60 & 6.25 & 87.96 \\
     & \textbf{CDG} & \textbf{33.07} & \textbf{31.96} & \underline{6.26} & \textbf{92.61} \\
    \midrule
    \multirow{4}{*}{\textbf{FLUX.1}} & CFG & 38.55 & \underline{31.20} & 6.06 & \underline{90.31} \\
     & CADS & 38.73 & \textbf{31.21} & 6.01 & 90.05\\
     & ICG & \underline{37.44} & 31.15 & \underline{6.11} & 90.21 \\
     & \textbf{CDG} & \textbf{37.11} & \textbf{31.21} & \textbf{6.15} & \textbf{90.62} \\
    \midrule
    \multirow{2}{*}{\textbf{Qwen}} & CFG & 42.45 & 32.11 & \textbf{2.57} & 93.66 \\
     & \textbf{CDG} & \textbf{39.02} & \textbf{32.31} & 2.54 & \textbf{93.93} \\
    \bottomrule
    \end{tabular}
    \label{tab:main_results}
\end{table}

We compare CDG against baselines including CFG~\cite{ho2022classifier}, CADS~\cite{sadat2024cads}, ICG~\cite{sadat2024no}, PAG~\cite{ahn2024selfrectifying}, SEG~\cite{hong2024smoothed}, SFG~\cite{azarian2024sfg}, and DNP~\cite{desai2024dnp} under identical evaluation settings.

\textbf{Quantitative Results.}
As shown in \cref{tab:main_results}, CDG consistently improves over CFG on all four backbones, achieving best or near-best results in most categories.

\textbf{Cross-Model Analysis.} \cref{tab:main_results} reveals CDG's improvements are more pronounced on SD3/SD3.5 than on FLUX.1, consistent with their training paradigms: FLUX.1 employs \textit{Guidance Distillation}~\cite{flux2024}, reducing dependence on inference-time guidance. To further validate generalization, we include Qwen-Image, which uses special tokens (\texttt{<|im\_end|>}) instead of padding as context aggregators. CDG consistently improves over CFG on this architecture (\cref{tab:main_results,tab:advanced_skills_main}), confirming that stratified degradation generalizes beyond padding-token architectures to any model with content/context-aggregating token structure.

\textbf{Qualitative Results.}
\cref{fig:title_image} presents representative comparisons across three models, illustrating typical failure modes of CFG on compositional prompts. CFG struggles with text rendering (producing misspelled words), spatial-attribute binding (confusing positions or attributes), and complex interactions (generating semantically ambiguous compositions). In contrast, CDG consistently achieves accurate rendering, precise spatial-semantic alignment, and correct action semantics---improvements consistent with quantitative gains on compositional benchmarks (\cref{tab:advanced_skills_main}). Additional results are in Appendix~C.9.

\textbf{Benchmark Results.}
We evaluate CDG on GenAI-Bench. \cref{tab:advanced_skills_main} shows CDG consistently outperforms all baselines on SD3.5, with particularly strong gains on \textit{Differentiation} (+3.64) and \textit{Comparison} (+2.36)---tasks requiring subtle semantic contrasts where CDG's ``good vs.\ almost good'' paradigm excels. CDG also significantly outperforms structure-level methods (PAG, SEG) that lack semantic awareness. Full results are in Appendix~C.7.

\begin{table}[t!]
    \centering
    \caption{GenAI-Bench compositional reasoning results: Spatial Relation (Spatial), Comparison (Comp), Differentiation (Differ), and Universal (Univ).}
    \small
    \begin{tabular}{cccccc}
    \toprule
    \multicolumn{2}{c}{\textbf{Method}} & \textbf{Spatial}$\uparrow$ & \textbf{Comp}$\uparrow$ & \textbf{Differ}$\uparrow$ & \textbf{Univ}$\uparrow$ \\
    \midrule
    \multirow{6}{*}{\textbf{SD3}} & CFG & \underline{78.66} & \underline{74.08} & \underline{77.22} & \underline{70.74} \\
     & CADS & 78.55 & 74.05 & 76.98 & 70.11 \\
     & ICG & 78.09 & 73.10 & 75.89 & 69.39 \\
     & SEG & 72.26 & 69.51 & 70.46 & 65.52 \\
     & PAG & 69.65 & 66.40 & 67.09 & 64.80 \\
     & \textbf{CDG} & \textbf{79.84} & \textbf{74.86} & \textbf{78.26} & \textbf{71.53} \\
    \midrule
    \multirow{6}{*}{\textbf{SD3.5}} & CFG & \underline{79.66} & 73.70 & 75.10 & \underline{72.21} \\
     & CADS & 79.58 & 73.54 & 75.08 & 71.94 \\
     & ICG & 78.90 & \underline{74.13} & \underline{75.63} & 70.23 \\
     & SEG & 76.34 & 71.43 & 72.80 & 67.73 \\
     & PAG & 75.64 & 71.02 & 72.38 & 66.17 \\
     & \textbf{CDG} & \textbf{80.69} & \textbf{76.06} & \textbf{78.74} & \textbf{73.13} \\
    \midrule
    \multirow{4}{*}{\textbf{FLUX.1}} & CFG & \underline{77.47} & \underline{72.97} & \underline{75.39} & 71.13 \\
     & CADS & 77.42 & 72.58 & 74.87 & 70.81 \\
     & ICG & 77.07 & 72.83 & 74.50 & \underline{71.47} \\
     & \textbf{CDG} & \textbf{77.56} & \textbf{73.47} & \textbf{76.17} & \textbf{71.55} \\
    \midrule
    \multirow{2}{*}{\textbf{Qwen}} & CFG & 83.26 & 80.19 & 83.41 & 77.36 \\
     & \textbf{CDG} & \textbf{83.79} & \textbf{80.24} & \textbf{83.54} & \textbf{77.56} \\
    \bottomrule
    \end{tabular}
    \label{tab:advanced_skills_main}
\end{table}

\subsection{Ablation Study}
\label{ssec:ablation}

Our ablation study examines four aspects: (1) component necessity (\cref{sssec:components}), (2) hyperparameter configuration (\cref{sssec:hyperparams}), (3) mechanism validation (\cref{sssec:cfg_star}), and (4) computational efficiency (\cref{sssec:efficiency}). In \cref{sssec:components}, we use a fixed degradation budget ($R_{\text{deg}}=1.1$) to isolate the contribution of each component; \cref{ssec:sota} reports results at the default $R_{\text{deg}}=1.0$ identified in \cref{sssec:hyperparams}.

\subsubsection{Core Component Analysis}
\label{sssec:components}

We demonstrate that WPR-based ranking effectively captures semantic structure. Details of experimental setup are in Appendix~C.3.

\cref{tab:ablation_factorial} reveals that Stratified Degradation is the primary driver of CDG's effectiveness. Both stratified variants (rows 1--2) dramatically outperform all non-stratified variants (rows 3--5), with VQA improvements of +5.9--12.2 points and FID reductions of 0.9--16.8 points, confirming that treating content and context-aggregating tokens as separate degradation pools is crucial for precise semantic control.

Rows 1 and 2 show comparable performance between WPR-based and random ranking within the stratified framework (FID: 33.89 vs.\ 34.17), confirming that WPR serves as an analysis tool providing determinism and theoretical grounding for the $R_{\text{deg}}=1.0$ boundary, rather than a necessary component.

Within non-stratified variants, WPR-based ranking (row 3) significantly outperforms reverse ranking (row 4, FID 50.73) and random ranking (row 5, FID 47.02), validating that WPR correctly identifies semantically important tokens when ranking is applied.

\begin{table}[h]
\centering
\caption{Ablation study on CDG core components at fixed degradation budget ($R_{\text{deg}}=1.1$). `\ding{51}' and `\ding{55}' denote active and inactive components. Asterisk ($^{*}$) indicates reverse ranking (least important tokens).}
\small
\setlength{\tabcolsep}{1mm}
\begin{tabular}{cc|cccc}
\toprule
\multicolumn{2}{c|}{\textbf{Components}} & \multicolumn{4}{c}{\textbf{Metrics}} \\
\cmidrule(lr){1-2} \cmidrule(lr){3-6}
\textbf{Importance} & \textbf{Stratified} & \makecell{\textbf{FID}\\}$\downarrow$ & \makecell{\textbf{CLIP}\\\textbf{Score}}$\uparrow$ & \makecell{\textbf{Aesthetic}\\\textbf{Score}}$\uparrow$ & \makecell{\textbf{VQA}\\\textbf{Score}}$\uparrow$ \\
\midrule
\ding{51} & \ding{51} & \textbf{33.89} & \underline{31.98} & \textbf{5.68} & \underline{92.21} \\
\ding{55} & \ding{51} & \underline{34.17} & \textbf{32.02} & \textbf{5.68} & \textbf{92.27} \\
\midrule
\ding{51} & \ding{55} & 35.06 & 30.93 & 5.48 & 86.31 \\
\ding{51}$^{*}$ & \ding{55} & 50.73 & 29.86 & 5.22 & 80.10 \\
\ding{55} & \ding{55} & 47.02 & 30.39 & 5.21 & 83.55 \\
\bottomrule
\end{tabular}
\label{tab:ablation_factorial}
\end{table}

\subsubsection{Optimal Hyperparameter Analysis}
\label{sssec:hyperparams}

Having established that WPR effectively captures semantic structure (\cref{sssec:components}), we now analyze hyperparameter selection. This section demonstrates our hyperparameter selection process on the SD3 model, which guided the configurations reported in \cref{tab:main_results}.

We analyze two key hyperparameters: the intervention block $\lambda_{\text{block}}$ and the unified Degradation Ratio $R_{\text{deg}} \in [0, 2]$ introduced in \cref{sec:method}, which maps to per-type degradation ratios ($r_{\text{content}}, r_{\text{CtxAgg}}$) via \cref{eq:unified_ratio}.

\begin{figure}[h]
    \centering
    \includegraphics[width=0.47\linewidth]{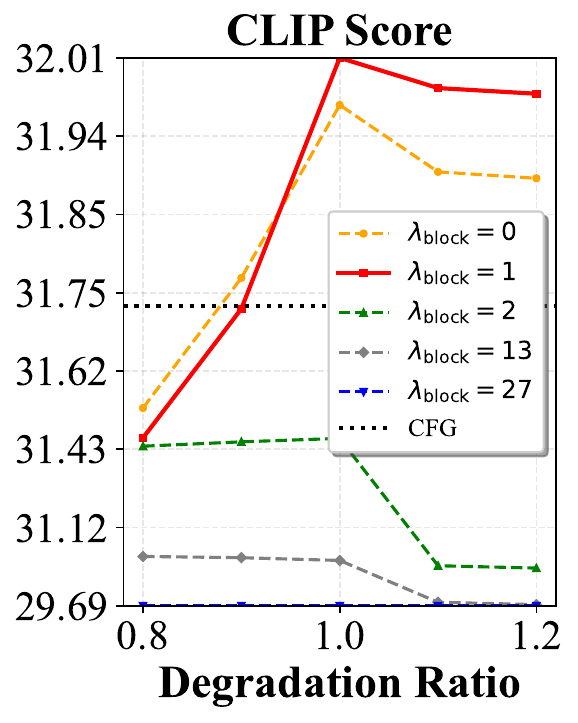} 
    \includegraphics[width=0.47\linewidth]{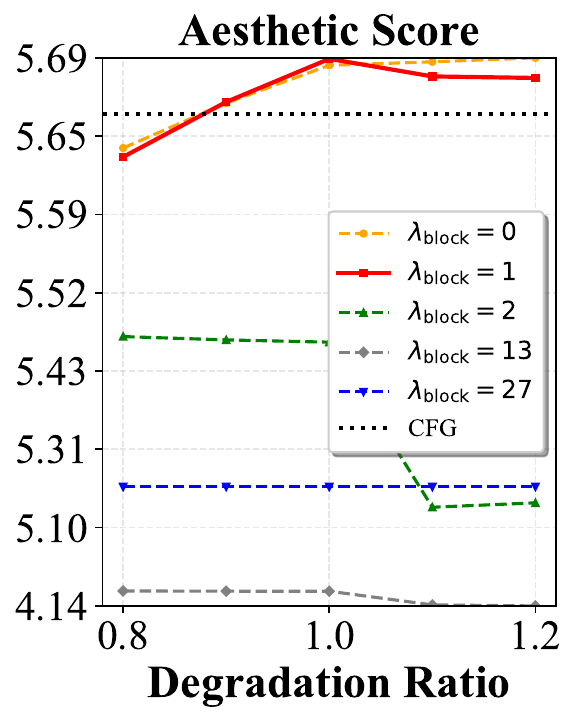} 
    \caption{
    \textbf{Hyperparameter analysis:} joint effect of intervention block ($\lambda_{\text{block}}$) and Degradation Ratio ($R_{\text{deg}}$) on SD3.
    }
    \label{fig:ablation_hyperparams}
\end{figure}

As shown in \cref{fig:ablation_hyperparams}, metrics exhibit an asymmetric response around $R_{\text{deg}}=1.0$, consistent with the content/context-aggregating dichotomy. The steep slope in $[0, 1.0]$ (removing content tokens) transitions to a gentler slope in $[1.0, 2.0]$ (removing context-aggregating tokens), with the latter region exhibiting relative stability for fine-grained style control. Experiments confirm $\lambda_{\text{block}}=1$ provides the most robust performance.
We adopt $R_{\text{deg}}=1.0$ as the default due to: (1) multi-metric balance, (2) cross-model applicability (detailed in Appendix~C.4), and (3) computational efficiency (all content tokens are degraded at this boundary, bypassing WPR entirely).

To further illustrate the impact of degradation ratio adjustments, Appendix~C.5 demonstrates qualitative results of adjusting degradation ratio near optimal parameters.

\subsubsection{CFG* Validation Experiment}
\label{sssec:cfg_star}

To validate the semantic properties of our constructed $\boldsymbol{c}_{\text{deg}}$, we design a CFG* experiment where we replace the positive prompt $\boldsymbol{c}$ in \cref{eq:cfg} with $\boldsymbol{c}_{\text{deg}}$, effectively using the degraded condition to guide generation. This allows us to directly probe what semantic information remains in $\boldsymbol{c}_{\text{deg}}$ at different degradation levels. Detailed formulation and additional qualitative results are provided in Appendix~C.6.

\begin{figure}[h]
    \centering
    \includegraphics[width=0.48\textwidth]{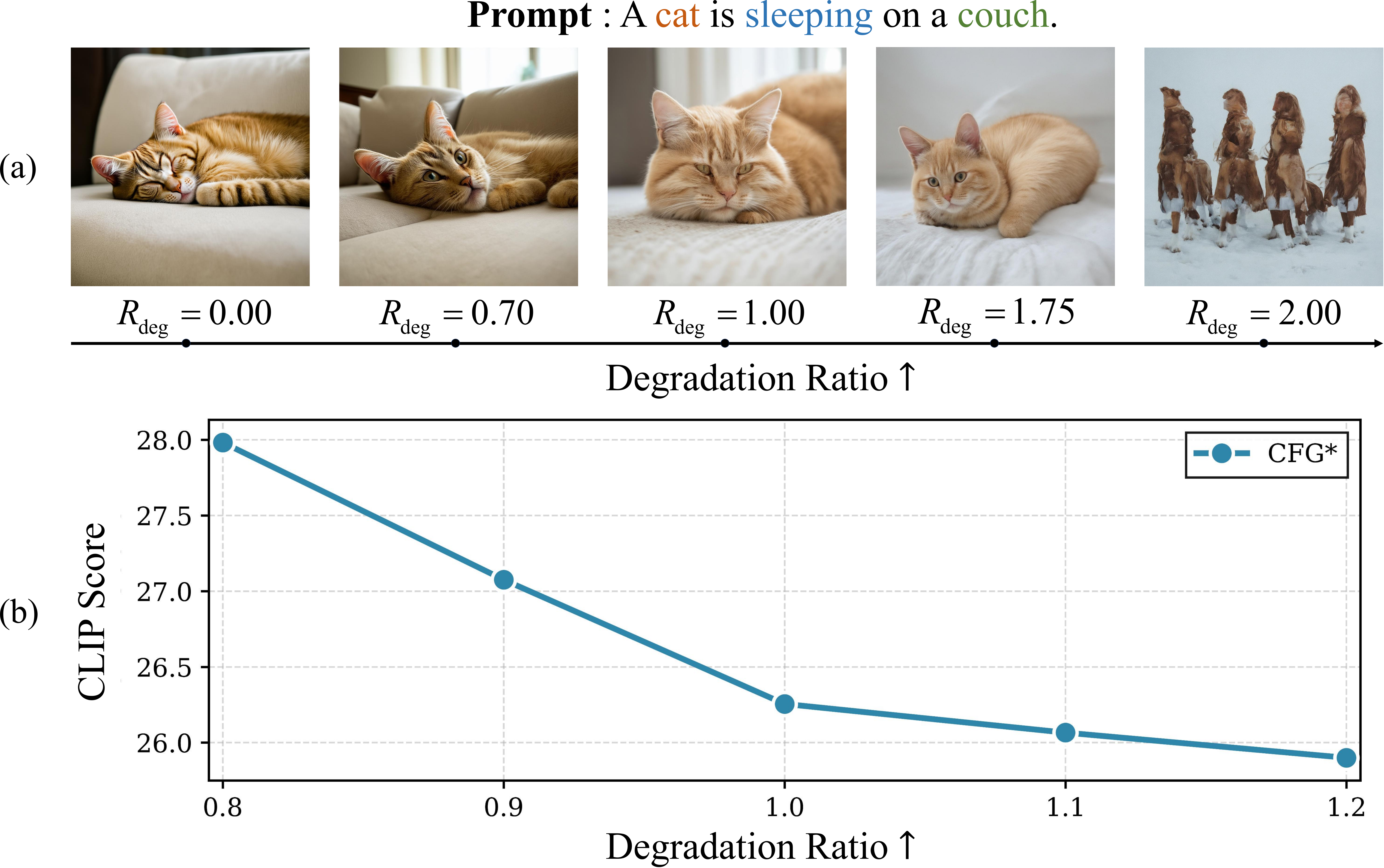}
    \caption{
        (a) Semantic degradation sequence under CFG* as $R_{\text{deg}}$ increases from 0.00 to 2.00. (b) Quantitative analysis.
    }
    \label{fig:cfg_star_analysis}
\end{figure}

\textbf{Qualitative Analysis.} As shown in \cref{fig:cfg_star_analysis}(a), under a fixed prompt, the generated results progressively lose semantic information as $R_{\text{deg}}$ increases from 0.00 to 2.00. In the range [0, 1], specific details such as ``sleeping'' and ``sofa'' are lost, while in [1, 2], the main subject ``cat'' disappears.

\textbf{Quantitative Analysis.} As shown in \cref{fig:cfg_star_analysis}(b), the CLIP Score exhibits monotonic decline with a noticeable slope change near $R_{\text{deg}} \approx 1.0$. This inflection point aligns with the content/context-aggregating boundary revealed by WPR (\cref{fig:token_importance_visualization}) and corresponds with the qualitative analysis, supporting our hypothesis that these two token types encode different semantic granularities. Content token removal ($R_{\text{deg}} \in [0, 1.0]$) causes steep decline due to loss of specific semantics, while context-aggregating token removal ($R_{\text{deg}} > 1.0$) shows gentler decline as only global context degrades. Together with the asymmetric pattern in \cref{fig:ablation_hyperparams}, these observations provide converging evidence for the dichotomy-driven design.

\subsubsection{Computational Efficiency Analysis}
\label{sssec:efficiency}

CDG's efficiency is critical. As shown in \cref{tab:ablation_efficiency}, a naive per-step recomputation incurs substantial overhead (+47.2\%) with negligible performance difference compared to one-time computation. In contrast, our one-time computation strategy introduces only minimal overhead (+3.6\%). Note that we use $R_{\text{deg}}=1.1$ for this analysis to evaluate WPR's computational cost; at $R_{\text{deg}}=1.0$, WPR computation is bypassed entirely (all content tokens are degraded), making overhead assessment infeasible. Crucially, at our default $R_{\text{deg}}=1.0$ setting, the strategy becomes a simple ``replace all content tokens'' operation, achieving near-zero overhead.

\begin{table}[h]
    \caption{Efficiency comparison of CDG implementation variants (one-time vs.\ per-step WPR computation) on SD3 with $R_{\text{deg}}=1.1$, $\lambda_{\text{block}}=1$.}
    \small
    \centering
    {%
    \small
    \setlength{\tabcolsep}{4pt}
    \renewcommand{\arraystretch}{0.95}
    \begin{tabular}{lccc}
    \toprule
    \textbf{Method} & \textbf{Time (s)} & \textbf{Aesthetic} & \textbf{VQA} \\
    \midrule
    CFG (Baseline) & 5.456 & 5.66 & 91.44 \\
    \midrule
    \makecell[l]{CDG (Per-Step)} & 8.031 (+47.2\%) & 5.68 & 92.33 \\
    \textbf{CDG} & \textbf{5.655 (+3.6\%)} & 5.68 & 92.21 \\
    \bottomrule
    \end{tabular}%
    }
    \label{tab:ablation_efficiency}
\end{table}

\subsection{Modularity and Applications}
\label{ssec:applications}

CDG is a plug-and-play module that operates directly on text embeddings, making it naturally compatible with existing methods and downstream tasks. We demonstrate its extensibility across three scenarios: (1) combination with orthogonal methods like PAG~\cite{ahn2024selfrectifying}, (2) image-to-image translation, and (3) ControlNet-based controllable generation, with comprehensive results in Appendix~C.9.

Appendix~B illustrates the CDG pipeline and key code, demonstrating the high extensibility of our method.

\section{Conclusion}
CFG's reliance on the semantically vacuous null prompt $\emptyset$ produces geometrically entangled guidance signals, limiting compositional accuracy. Our Condition-Degradation Guidance (CDG) addresses this by constructing a semantically degraded condition $\boldsymbol{c}_{\text{deg}}$ through attention-based analysis, reframing guidance as ``good vs.\ almost good'' discrimination. This design enables common-mode rejection, synthesizing guidance signals with superior orthogonality to the denoising manifold. Validated on state-of-the-art models, CDG consistently improves compositional reasoning and text-image alignment with negligible overhead. This work establishes a principle: adaptive, semantically-aware negative synthesis is essential for precise semantic control in conditional diffusion models.

\section*{Acknowledgements}
We sincerely thank four anonymous reviewers for their valuable and constructive feedback, which has greatly improved our work. This work was supported by the following grants: the National Natural Science Foundation of China (Grant No.\ 12471401, 12401419).
{
    \small
    \bibliographystyle{ieeenat_fullname}
    \bibliography{main}
}

\clearpage
\maketitlesupplementary

\appendix

\section{Token Importance Analysis}
\label{appendix:analysis}

The method for calculating token importance described in this section is adapted from the Zero-TPrune model proposed by Wang et al. It is important to note that Zero-TPrune was originally designed for the \textbf{visual domain}, where it analyzes the attention graph to calculate the importance of \textbf{image} patches (also treated as tokens) for pruning Vision Transformers. In this study, we migrate and apply this methodology from the visual domain to the \textbf{textual domain}, aiming to quantify the semantic importance of each text token within a natural language context.

The method consists of two main components:
\begin{enumerate}
    \item \textbf{Weighted PageRank (WPR) Algorithm}: Used to calculate token importance scores within a single attention head.
    \item \textbf{Multi-Head Score Fusion Strategy}: Combines a Variance-based Head Filter (VHF) and Emphasizing Informative Region (EIR) aggregation to fuse scores from multiple heads.
\end{enumerate}

\subsection{Single-Head Importance via Weighted PageRank}
\label{sec:wpr}

The WPR algorithm treats the attention matrix $A$ from a single attention head as the adjacency matrix of a weighted directed graph. It iteratively calculates the importance score of each node (i.e., token) in the graph. The importance of a token is determined by the weighted importance of other tokens that point to it. The calculation process is shown in Algorithm~\ref{alg:text_pagerank}.

\begin{algorithm}[h]
\caption{PageRank for Token Importance Scoring}
\label{alg:text_pagerank}
\begin{algorithmic}[1]
\REQUIRE Attention matrix $A$, token length $n$
\REQUIRE Convergence threshold $\epsilon$, maximum iterations $T$
\ENSURE Token importance scores $s$

\STATE $A = \text{row\_norm}(A)$
\STATE Initialize uniform importance scores $s^{(0)} = \frac{1}{n} \mathbf{1}_{n}$
\FOR{iteration $t = 1$ to $T$}
    \STATE $s^{prev} = s^{(t-1)}$
    \STATE $s^{(t)} = A^T s^{prev}$
    \STATE $s^{(t)} = s^{(t)} / \|s^{(t)}\|_1$
    \IF{$\|s^{(t)} - s^{prev}\|_1 < \epsilon$}
        \STATE \textbf{break}
    \ENDIF
\ENDFOR
\RETURN $s^{(t)}$
\end{algorithmic}
\end{algorithm}

\subsection{Multi-Head Score Fusion}
\label{sec:fusion}

For a multi-head attention mechanism, after obtaining an independent score distribution from each head via the WPR algorithm, these scores must be fused into a final score. A simple averaging of scores is not optimal, as different heads may focus on different semantic patterns. The method employs a fusion strategy that combines a Variance-based Head Filter (VHF) and Emphasizing Informative Region (EIR) aggregation.

\begin{enumerate}
    \item \textbf{Variance-based Head Filter (VHF):} This technique is used to identify and filter out "bad" heads that provide uninformative or misleading score distributions. The procedure involves calculating the variance, $\mathrm{Var}_h$, of the importance score distribution for each head. Only heads whose variance falls within a predefined threshold range [$v_{\min}, v_{\max}$] are retained; the rest are discarded from subsequent calculations.

    \item \textbf{Emphasizing Informative Region (EIR):} For the valid heads filtered by VHF, this method uses a "root-mean of sum of squares" approach for score aggregation. Compared to a direct average, the squaring operation amplifies high scores from any single head. This ensures that a token's high importance is not diluted in the final score, even if it is deemed critical by only a few "specialist" heads.
\end{enumerate}

\textbf{Final Score Calculation}

Combining the VHF and EIR strategies, the final importance score $s^{(l)}(x_i)$ for the $i$-th token in the $l$-th layer is calculated as follows:

$$s^{(l)}\left(x_{i}\right)=\sqrt{\frac{\sum_{h=1}^{N_{h}} s^{(h, l)}\left(x_{i}\right)^{2} \cdot \eta\left(v_{\min } \leq \mathrm{Var}_{h} \leq v_{\max }\right)}{\sum_{h=1}^{N_{h}} \eta\left(v_{\min } \leq \mathrm{Var}_{h} \leq v_{\max }\right)}}$$

Where:
\begin{itemize}
    \item $s^{(h, l)}(x_i)$ is the importance score of token $x_i$ from head $h$ in layer $l$.
    \item $\mathrm{Var}_h$ is the variance of the score distribution for head $h$.
    \item $\eta(\cdot)$ is an indicator function that implements the VHF. The function returns 1 if the condition $v_{\min} \leq \mathrm{Var}_h \leq v_{\max}$ is met, and 0 otherwise.
    \item $N_h$ is the total number of heads in the current layer.
\end{itemize}

\subsection{Cross attention}

A natural baseline for computing token importance is to leverage the cross-attention mechanism between text and image tokens. In diffusion models, the cross-attention matrix $C \in \mathbb{R}^{N_{\text{img}} \times N_{\text{text}}}$ is computed where image tokens serve as queries and text tokens serve as keys. For each text token $i$, its attention weights across all image positions form the $i$-th column of $C$. These weights are typically reshaped to the spatial dimensions of the latent to produce saliency maps.

Following this intuition, one might compute the importance score of text token $i$ by simply summing its corresponding column:
\begin{equation}
s_i^{\text{cross}} = \sum_{j=1}^{N_{\text{img}}} C_{j,i}
\end{equation}

\begin{figure*}[h]
\centering
\includegraphics[width=0.95\linewidth]{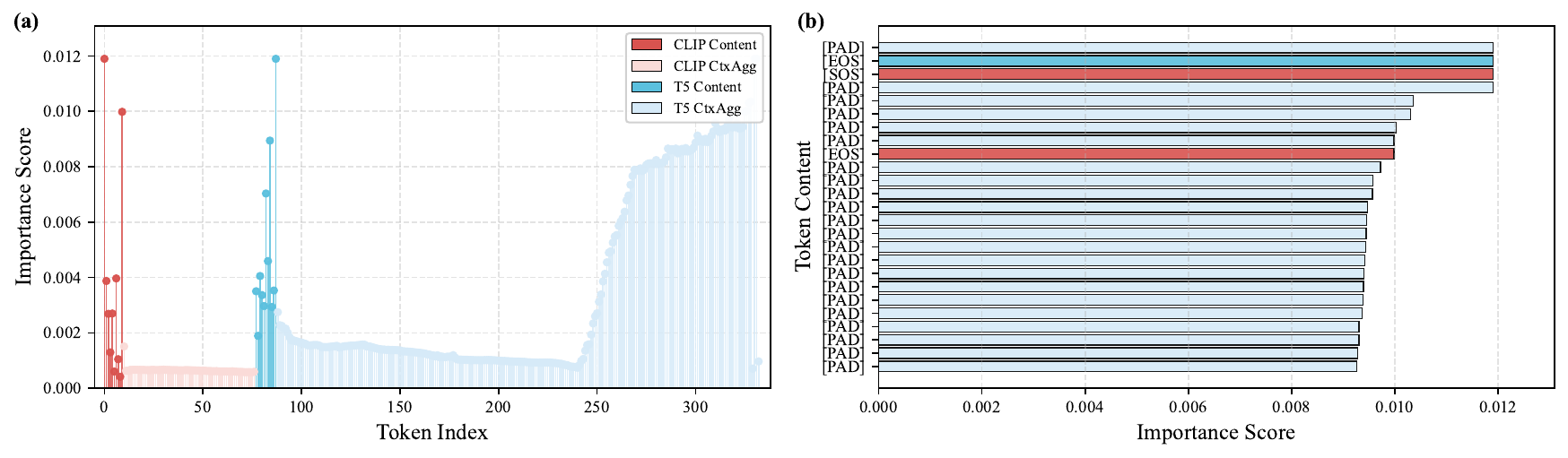}
\caption{Token importance scores computed via cross-attention summation for the prompt ``A man is cooking, Minecraft Style.'' The method incorrectly assigns highest importance to padding tokens rather than semantic content tokens.}
\label{fig:cross_attn_failure}
\end{figure*}

However, this approach yields counterintuitive results. Consider the prompt ``A man is cooking, Minecraft Style.'' As shown in Figure~\ref{fig:cross_attn_failure}, the cross-attention-based importance scores assign the highest values predominantly to padding tokens, rather than to semantically meaningful tokens like ``man'', ``cooking'', or ``Minecraft''. This paradoxical behavior contradicts our intuition and renders cross-attention unsuitable for token importance estimation.

This failure motivates our graph-based approach using self-attention and Weighted PageRank, which properly captures token relationships and avoids the padding token bias inherent in cross-attention aggregation.

\section{Implementation Details}
\label{appendix:implementation}

This section provides the complete algorithmic and implementation details of our Condition-Degradation Guidance (CDG). We present: (1) the full sampling algorithm that integrates CDG into the diffusion denoising loop, and (2) the core PyTorch implementation for token degradation.

\subsection{Algorithm: CDG Sampling}

Algorithm~\ref{alg:cdg_sampling} details the complete CDG sampling procedure, which augments the standard diffusion sampling loop with our semantic degradation mechanism. The algorithm consists of three key components:

\paragraph{CalculateImportance} (lines 2--7) computes token importance scores using the Weighted PageRank (WPR) algorithm on the self-attention graph. It extracts the attention affinity matrix $\boldsymbol{A}$ from transformer block $\lambda_{\text{block}}$, applies WPR and multi-head fusion, and returns sorted token indices $\boldsymbol{i}_{\text{sorted}}$ in descending order of importance. For efficiency, this is computed only once at $t=T$ and cached for all subsequent steps.

\paragraph{ProcessToken} (lines 9--11) constructs the degraded condition $\boldsymbol{c}_{\text{deg}}$ at each timestep. Using the cached importance ranking and degradation ratios $r_{\text{deg}} = \{r_{\text{content}}, r_{\text{CtxAgg}}\}$, it builds a binary mask $\boldsymbol{m}$ that marks less important tokens for replacement, then performs masked interpolation: $\boldsymbol{c}_{\text{deg}} \leftarrow \boldsymbol{m} \odot \boldsymbol{c} + (1 - \boldsymbol{m}) \odot \boldsymbol{c}_{\emptyset}$.

\paragraph{Main Sampling Loop} (lines 13--22) performs guided denoising for $T$ steps. At each timestep $t$, it computes two noise predictions—$\boldsymbol{\epsilon}_{\text{cond}}$ with full condition $\boldsymbol{c}$ and $\boldsymbol{\epsilon}_{\text{deg}}$ with degraded condition $\boldsymbol{c}_{\text{deg}}$—then combines them via the CDG formula: $\hat{\boldsymbol{\epsilon}} \leftarrow \boldsymbol{\epsilon}_{\text{cond}} + (w-1) \cdot (\boldsymbol{\epsilon}_{\text{cond}} - \boldsymbol{\epsilon}_{\text{deg}})$, where $w$ is the guidance scale.

\begin{algorithm}[h]
    \caption{Sampling with Condition-Degradation Guidance (CDG)}
    \label{alg:cdg_sampling}
    \begin{algorithmic}[1]
    
    \REQUIRE Initial noise $\boldsymbol{x}_T \sim \mathcal{N}(0, \boldsymbol{I})$, text embedding $\boldsymbol{c}$, empty embedding $\boldsymbol{c}_{\emptyset}$, guidance scale $w$, denoising steps $T$, denoiser $D_\theta$, transformer block index $\lambda_{\text{block}}$, degradation ratio $r_{\text{deg}} = \{r_{\text{content}}, r_{\text{CtxAgg}}\}$.
    \ENSURE Generated image $\boldsymbol{x}_0$.
    
    \vspace{0.5em}
    \STATE \textbf{CalculateImportance}($\boldsymbol{c}, D_\theta, \lambda_{\text{block}}$):
    \STATE \hspace*{\algorithmicindent} $\boldsymbol{Q}, \boldsymbol{K} \leftarrow \textsc{ExtractQK}(D_\theta, \lambda_{\text{block}}, \boldsymbol{c})$
    \STATE \hspace*{\algorithmicindent} $\boldsymbol{A} \leftarrow \boldsymbol{Q} \boldsymbol{K}^T$
    \STATE \hspace*{\algorithmicindent} $\boldsymbol{s}_{\text{heads}} \leftarrow \textsc{WPR}(\boldsymbol{A})$
    \STATE \hspace*{\algorithmicindent} $\boldsymbol{s} \leftarrow \textsc{ScoreFusion}(\boldsymbol{s}_{\text{heads}})$
    \STATE \hspace*{\algorithmicindent} $\boldsymbol{i}_{\text{sorted}} \leftarrow \textsc{ArgSort}(\boldsymbol{s})$
    \RETURN $\boldsymbol{i}_{\text{sorted}}$
    
    \vspace{0.5em}
    \STATE \textbf{ProcessToken}($\boldsymbol{c}, \boldsymbol{c}_{\emptyset}, \boldsymbol{i}_{\text{sorted}}, r_{\text{deg}}$):
    \STATE \hspace*{\algorithmicindent} $\boldsymbol{m} \leftarrow \mathrm{BuildMask}(\boldsymbol{i}_{\text{sorted}}, r_{\text{deg}})$
    \STATE \hspace*{\algorithmicindent} $\boldsymbol{c}_{\text{deg}} \leftarrow \boldsymbol{m} \odot \boldsymbol{c} + (1 - \boldsymbol{m}) \odot \boldsymbol{c}_{\emptyset}$
    \RETURN $\boldsymbol{c}_{\text{deg}}$
    
    \vspace{0.5em}
    \STATE \textbf{// Main sampling process}

    \FOR{$t = T, T-1, \ldots, 1$}
        \IF{$t=T$}
        \STATE $\boldsymbol{i}_{\text{sorted}} \leftarrow \textsc{CalculateImportance}(\boldsymbol{c}, D_\theta, \lambda_{\text{block}})$
        \ENDIF
        \STATE $\boldsymbol{c}_{\text{deg}} \leftarrow \textsc{ProcessToken}(\boldsymbol{c}, \boldsymbol{c}_{\emptyset}, \boldsymbol{i}_{\text{sorted}}, r_{\text{deg}})$ 
        \STATE $\boldsymbol{\epsilon}_{\text{cond}} \leftarrow D_\theta(\boldsymbol{x}_t, t, \boldsymbol{c})$
        \STATE $\boldsymbol{\epsilon}_{\text{deg}} \leftarrow D_\theta(\boldsymbol{x}_t, t, \boldsymbol{c}_{\text{deg}})$
        \STATE $\hat{\boldsymbol{\epsilon}} \leftarrow \boldsymbol{\epsilon}_{\text{cond}} + (w-1) \cdot (\boldsymbol{\epsilon}_{\text{cond}} - \boldsymbol{\epsilon}_{\text{deg}})$
        \STATE $\boldsymbol{x}_{t-1} \leftarrow \textsc{SchedulerStep}(\boldsymbol{x}_t, \hat{\boldsymbol{\epsilon}}, t)$
    \ENDFOR
    
    \RETURN $\boldsymbol{x}_0$
    \end{algorithmic}
\end{algorithm}

\subsection{Core Implementation}

The code listing below presents the PyTorch implementation of the token degradation logic (corresponding to the \texttt{ProcessToken} function in Algorithm~\ref{alg:cdg_sampling}). The implementation includes three key optimizations: (1) early exit when no degradation is needed (lines 107--108), (2) importance caching controlled by \texttt{all\_use\_first\_step\_importance} flag (lines 111--119), and (3) efficient masked interpolation via broadcasting (lines 128--130). This compact implementation demonstrates how CDG integrates seamlessly into existing diffusion pipelines with minimal code overhead.

\begin{lstlisting}[language=Python]
def __call__(self, positive_encoder_hidden_states, negative_encoder_hidden_states):
    """Process encoder hidden states based on importance and degrade ratios"""
    
    degrade_ratio = self.process_params["degrade_ratio"]
    
    # Optimization: Skip importance calculation if not needed
    if degrade_ratio in [{"content": 1, "CtxAgg": 0}]:
        sorted_indices = [i for i in range(len(positive_encoder_hidden_states))]
    
    # Use cached importance from first step (if enabled)
    elif (self.process_params.get("all_use_first_step_importance") 
          and self.first_step_importance is not None):
        sorted_indices = self.first_step_importance
    
    # Calculate importance (first step or every step depending on config)
    else:
        sorted_indices, scores = self.importance_calculator(positive_encoder_hidden_states, negative_encoder_hidden_states)
        if self.process_params.get("all_use_first_step_importance"):
            self.first_step_importance = sorted_indices  # Cache for later
    
    # Generate keep mask based on importance and ratios
    degrade_mask, degraded_indices = self.get_degrade_mask(
        sorted_indices,  
        ratio=degrade_ratio
    )
    
    # Interpolate: degraded = (1 - degrade_mask) * positive + degrade_mask * negative
    degrade_mask = degrade_mask.unsqueeze(0).unsqueeze(-1)  # Expand dimensions
    result = ((1 - degrade_mask) * positive_encoder_hidden_states + 
              degrade_mask * negative_encoder_hidden_states)
    
    return result
\end{lstlisting}
\section{Experimental Details and Supplementary Results}
\label{appendix:experiments}

This section provides comprehensive details for all experimental components referenced in the main paper, including evaluation metrics implementation, optimal hyperparameters for different models, and additional experimental results.

\subsection{Evaluation Metrics Implementation}
\label{appendix:metrics}

This section provides detailed implementation specifications for all evaluation metrics used in our experiments to ensure full reproducibility.

\subsubsection{Fréchet Inception Distance (FID).}
FID is a standard metric for assessing the quality of generated images by measuring the distributional distance to real images. To eliminate ambiguity, we specify that all our FID calculations are performed using the official implementation within the \textbf{\texttt{torchmetrics}} library. Following standard practice, we extract the 2048-dimensional features from the final pooling layer of the pre-trained Inception-v3 network.

\subsubsection{CLIP Score.}
The CLIP Score evaluates the semantic consistency between a generated image and its corresponding text prompt. Different CLIP model variants can yield varying scores. To ensure fair and reproducible comparisons, we consistently use the \texttt{openai/clip-vit-base-patch32} model as the backbone for all CLIP Score calculations.

\subsubsection{Aesthetic Score.}
The Aesthetic Score is a metric designed to computationally estimate the visual appeal of an image, simulating human perception of beauty. The score is generated by a dedicated predictor model, which typically consists of a regressor built upon a powerful, pre-trained vision foundation model like a CLIP image encoder. This predictor is trained on a large-scale dataset where images are paired with aesthetic ratings collected from human volunteers. This process allows the model to learn the visual features that correlate with positive human aesthetic judgment. In our work, we employ the model provided within the \texttt{aesthetic\_predictor\_v2\_5} library. A higher score from this model indicates a higher predicted visual appeal.

\subsubsection{VQA Score.}

The VQA (Visual Question Answering) Score assesses the factual alignment between a prompt and a generated image by framing the evaluation as a comprehension task. The process involves three steps: First, the descriptive prompt is programmatically converted into a closed-form (yes/no) question. Second, a pre-trained VQA model is presented with both the generated image and this question. Finally, the VQA Score is determined by the model's output probability for the affirmative answer, "Yes". A high score signifies high confidence from the VQA model that the image accurately depicts the content specified in the original prompt. For this evaluation, our implementation is based on the \texttt{t2v\_metrics} framework, and we utilize the \texttt{clip-flant5-xxl} model as the VQA engine. The question template is fixed as ``Does this figure show `[prompt]'? Please answer yes or no.''

\subsection{Dataset}

In this study, we employed two core benchmark datasets for model evaluation: the COCO 2017 validation set and GenAI Bench.

\subsubsection{COCO 2017 Validation Set}
This is an authoritative dataset widely used in the field of computer vision. It contains 5,000 images, each associated with five textual descriptions, totaling 25,000 captions. For consistency in our experiments, we uniformly selected the first caption for each image as the prompt for the generation task.

\subsubsection{GenAI Bench}
This is a novel dataset focused on evaluating the compositional reasoning capabilities of models. Its core consists of 1,600 high-quality text prompts sourced from the real-world workflows of professional designers. The dataset is specifically designed to test the performance of AI models in understanding and executing text-to-vision tasks that involve complex logic and compositional relationships, such as multiple objects, fine-grained attributes, spatial relations, and advanced reasoning.

\subsubsection{Rationale for Selection}
The rationale for selecting these two datasets is that they collectively form a \textbf{complementary and comprehensive evaluation framework}.
On one hand, the descriptive prompts from COCO 2017, which primarily depict common objects and general scenes, are ideal for measuring a model's \textbf{foundational capabilities and generalizability} in generating \textbf{common objects and conventional scenes}.
On the other hand, GenAI Bench presents a more rigorous challenge; its complex, compositional prompts allow for an in-depth probe of a model's \textbf{upper limits in advanced semantic understanding, logical reasoning, and precise detail control}.
By combining these datasets, our study can not only assess the model's baseline performance on conventional tasks but also conduct a more nuanced analysis of its strengths and weaknesses in handling complex, fine-grained instructions, thereby enabling a more holistic and multi-faceted evaluation of its overall capabilities.

\subsection{Optimal Hyperparameters for Different Models}
\label{appendix:hyperparams}

\begin{table}[ht!]
    \small
    \setlength{\tabcolsep}{2pt}
    \centering
    \begin{tabular}{clccccccc}
    \toprule
    \textbf{Model} & \textbf{Approach} & \textbf{$w$} & \textbf{$T$} & \textbf{$s$} & \textbf{seed} & $\sigma$ & \textbf{$\lambda_{\text{block}}$} & \textbf{$r_{\text{content}}|r_{\text{CtxAgg}}$} \\
    \midrule
    \multirow{6}{*}{\textbf{SD3}} 
    & CFG & 7 & 28 & - & - & - & - & - \\
    & CADS & 7 & 28 & 0.07 & - & - & - & - \\
    & ICG & 7 & 28 & - & 42 & - & - & - \\
    & PAG & 3 & 28 & - & - & - & 13 & - \\
    & SEG & 3 & 28 & - & - & 5 & 13 & - \\
    & CDG  & 7 & 28 & - & - & - & 1 & 1.0$|$0.0 \\
    \midrule
    \multirow{6}{*}{\textbf{SD3.5}} 
    & CFG & 3.5 & 28 & - & - & - & - & - \\
    & CADS & 3.5 & 28 & 0.07 & - & - & - & - \\
    & ICG & 3.5 & 28 & - & 42 & - & - & - \\
    & PAG & 3 & 28 & - & - & - & 13 & - \\
    & SEG & 3 & 28 & - & - & 5 & 13 & - \\
    & CDG  & 3.5 & 28 & - & - &- & 2 & 1.0$|$0.0 \\
    \midrule
    \multirow{4}{*}{\textbf{FLUX.1}} 
    & CFG & 1.5 & 28 & - & - & - & - & - \\
    & CADS & 1.5 & 28 & 0.07 & - & - & - & - \\
    & ICG & 1.5 & 28 & - & 42 & - & - & - \\
    & \textbf{CDG} & 1.5 & 28 & - &- & - & 1 & 1.0$|$0.0 \\
    \bottomrule
    \end{tabular}
    \caption{Optimal hyperparameters for different models. (``-'' denotes not used)}
    \label{tab:hyperparams}
\end{table}

This section details the parameters used for the CFG, CADS, ICG, and CDG methods across different models, as summarized in Table~\ref{tab:hyperparams}. The guidance scale ($w$) and the number of steps ($T$) were set based on default values or official examples. The parameter $s$ for CADS, which controls the intensity of guidance noise, was adopted from the value used for DiT models as specified in its paper's appendix. The seed for ICG (set to 42) initializes a random number generator used to select a random token ID for each prompt in the COCO dataset, ensuring reproducibility. For our CDG method, $\lambda_{\text{block}}$ denotes the layer index where degradation is applied, while $r_{\text{content}}$ and $r_{\text{CtxAgg}}$ represent the degradation ratios for content and context-aggregating tokens, respectively.

It is important to note that for the FLUX.1-dev model, the guidance scale $w$ corresponds to the \texttt{true\_cfg\_scale} parameter in its codebase, and we used an empty string for the negative prompt to implement the CFG method. Furthermore, while FLUX.1-dev is capable of generating high-quality, high-resolution images, its inference time is considerable. Due to computational constraints, we reduced the generation resolution for the FLUX model from 1024$\times$1024 to 512$\times$512 and decreased the number of inference steps from 50 to 28.

Additionally, it is important to clarify the interpretation of the guidance scale parameter $w$ in Table~\ref{tab:hyperparams}. For traditional guidance methods (CFG, CADS, ICG, CDG), $w$ directly represents the guidance strength. However, for PAG and SEG methods, which support a combined \texttt{cfg+pag} mode, the reported $w$ value corresponds to the PAG/SEG guidance scale, while the CFG scale is set to 1.0 (i.e., CFG is effectively disabled). This configuration allows PAG and SEG to operate independently without interference from CFG, which is the recommended setting in their respective implementations.

\subsection{Hyperparameters analysis}
\label{appendix:hyperparams_analysis}

While Section~6.3.2 of the main text presents a comprehensive hyperparameter analysis on SD3, establishing $R_{\text{deg}}=1.0$ as the optimal default, this section provides complementary ablation studies on SD3.5 and FLUX.1 to validate the cross-model generality of this choice. Our goal is to demonstrate that $R_{\text{deg}}=1.0$ serves as a robust initialization across diverse architectures, not merely a model-specific optimum.

\textbf{Experimental Setup.}
For each model, we first conducted a preliminary search over intervention block indices ($\lambda_{\text{block}}$) to identify the most promising configuration. Based on these initial experiments, we selected $\lambda_{\text{block}}=2$ for SD3.5 and $\lambda_{\text{block}}=1$ for FLUX.1. We then performed ratio ablation at these fixed indices, systematically varying $R_{\text{deg}}$ to observe metric responses. 

Due to computational constraints, these supplementary ablations were conducted on a subset of 500 images from the MS-COCO validation set, rather than the full 5,000 images used in the main SD3 analysis. This design choice is justified by two considerations: (1) the primary conclusion regarding $R_{\text{deg}}=1.0$ was already established through the comprehensive SD3 analysis on 5,000 images, and (2) the 500-image subset provides sufficient statistical power to observe trend patterns and validate consistency across models. The convergence of results across different sample sizes further supports the robustness of our findings.

\begin{figure*}[h]
    \centering
    \includegraphics[width=0.95\textwidth]{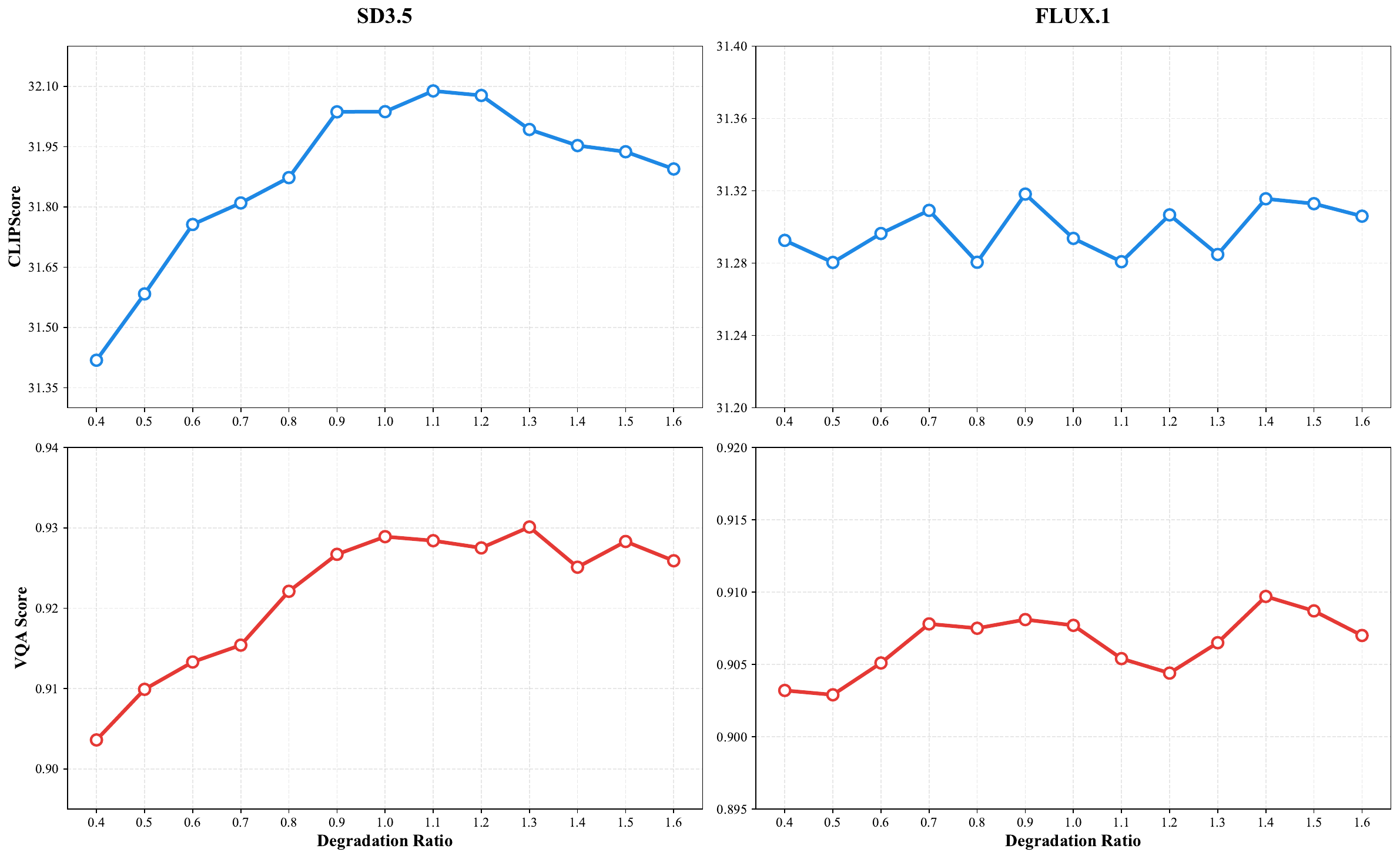}
    \caption{
    \textbf{Hyperparameter ablation on SD3.5 and FLUX.1:} Performance metrics (CLIP Score and VQA Score) as a function of Degradation Ratio $R_{\text{deg}}$. Top row: CLIP Score; bottom row: VQA Score. Left column: SD3.5 results; right column: FLUX.1 results. For SD3.5, both metrics achieve their best performance around $R_{\text{deg}}=1.0$-$1.3$, exhibiting a stable plateau that is consistent with the SD3 analysis in the main text. For FLUX.1, metrics exhibit relative stability across the tested range, with $R_{\text{deg}}=1.0$ providing a reliable initialization point.
    }
    \label{fig:combined_ablation}
\end{figure*}

\textbf{Results and Analysis.}
Figure~\ref{fig:combined_ablation} presents the ablation results for both models across CLIP Score and VQA Score metrics. For SD3.5 (left column), the patterns closely mirror those observed for SD3 in the main text: both metrics achieve their best performance in the region around $R_{\text{deg}}=1.0$, with CLIP Score reaching its maximum near $R_{\text{deg}}=1.1$ and VQA Score achieving optimal values around $R_{\text{deg}}=1.0$-$1.3$. Notably, the performance plateau in the $[1.0, 1.3]$ range demonstrates robustness to small variations in context-aggregating token degradation. The asymmetric response pattern—steeper sensitivity in $[0, 1.0]$ (content token removal) transitioning to gentler slopes in $[1.0, 2.0]$ (context-aggregating token removal)—confirms the content/context-aggregating dichotomy holds across SD3 variants.

For FLUX.1 (right column), the curves exhibit notably flatter profiles with reduced sensitivity to $R_{\text{deg}}$ variations. This behavior aligns with FLUX.1's training paradigm: as discussed in Section~6.2, FLUX.1 employs \textit{Guidance Distillation}, making it inherently less dependent on inference-time guidance mechanisms. Consequently, the model shows more stable performance across different degradation ratios. Despite this relative flatness, $R_{\text{deg}}=1.0$ remains a sensible initialization choice, as it falls within the stable, high-performing region for both metrics and maintains consistency with the SD3/SD3.5 configuration.

\textbf{Conclusion.}
These cross-model ablations validate the generality of $R_{\text{deg}}=1.0$ as a default initialization. While SD3.5 demonstrates the same optimal behavior as SD3, FLUX.1's relative insensitivity to ratio variations further supports the robustness of this choice: even when the model is less guidance-dependent, $R_{\text{deg}}=1.0$ provides reliable performance without requiring extensive per-model tuning. This cross-architectural consistency, combined with the computational efficiency benefits at $R_{\text{deg}}=1.0$ (as discussed in Section~6.3.4), establishes $R_{\text{deg}}=1.0$ as a principled default for CDG across diverse diffusion model architectures.

\subsection{Visual Examples of Varying Degradation Ratios}

\begin{figure}[h]
\centering
\includegraphics[width=\columnwidth]{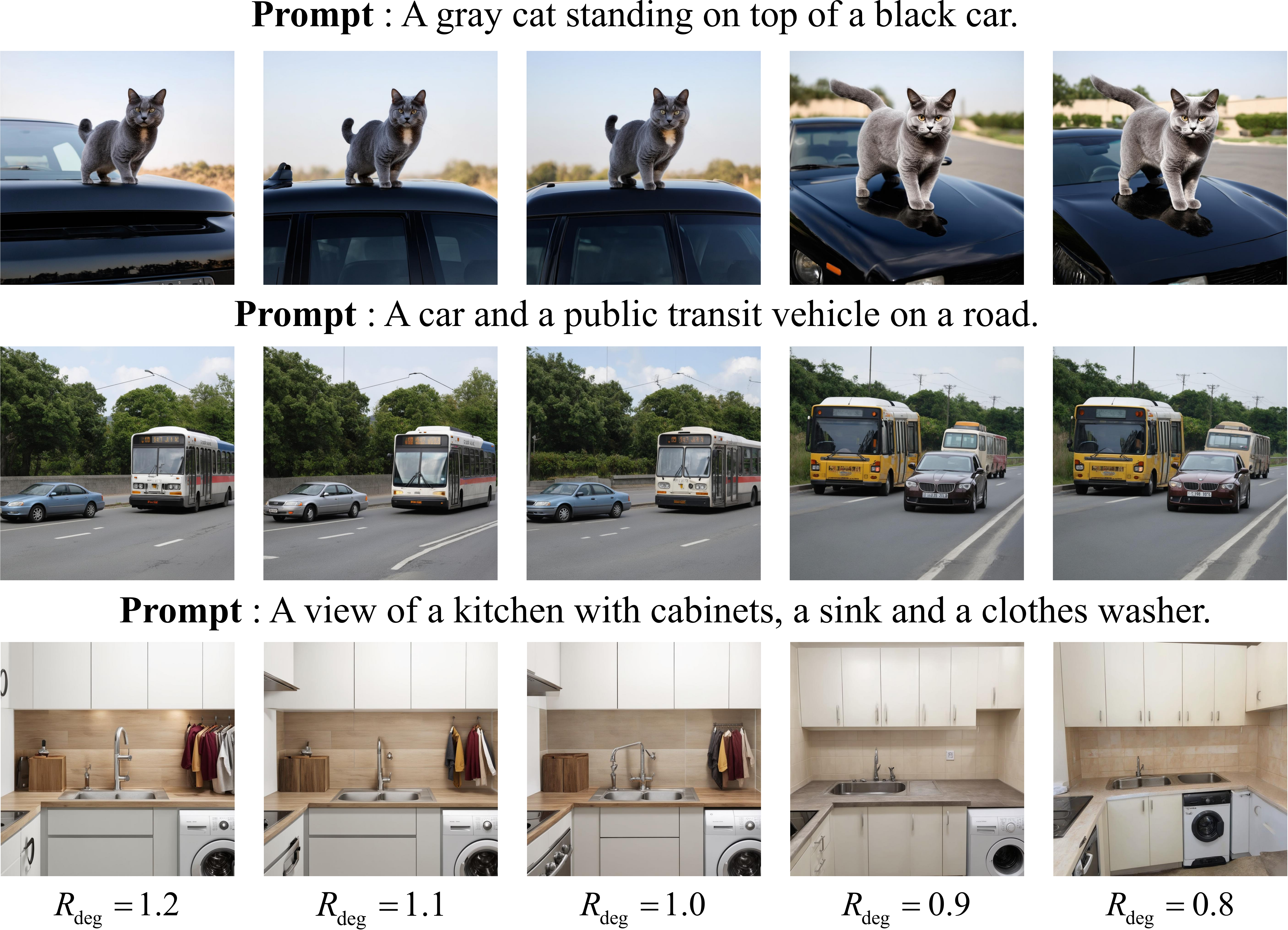}
\caption{Visual comparison of results generated by varying the degradation ratio $R_{\text{deg}}$.}
\label{fig:appendix_r_deg}
\end{figure}

Figure~\ref{fig:appendix_r_deg} visually demonstrates the impact of the unified degradation ratio $R_{\text{deg}}$ on the generated results.
It can be observed that when fine-tuning around the default configuration of $R_{\text{deg}}=1.0$, the model consistently generates images that adhere to the core elements of the prompt, yet they exhibit diverse variations in visual features such as composition and perspective.
This indicates that $R_{\text{deg}}$ not only ensures the alignment of key content but also serves as an effective control for users to flexibly adjust according to specific aesthetic preferences, allowing them to explore different generative styles.

\subsection{Additional Details for \texorpdfstring{CFG$^{*}$}{CFG*} Analysis}
\label{appendix:cfg_star_details}

This section provides detailed experimental setup and additional qualitative results for the CFG$^{*}$ analysis presented in Section~6.3.3 of the main text.

\subsubsection{Detailed Experimental Setup.}
\label{par:appendix_cfg_star_setup}

As discussed in the main text, CFG$^{*}$ replaces the positive prompt $\boldsymbol{c}$ in standard CFG (Eq.~(3)) with the degraded condition $\boldsymbol{c}_{\text{deg}}$. For completeness, we provide the explicit formulation and implementation details below.

\textbf{Standard CFG formulation:}
\begin{equation*}
    \hat{D}_\theta(\boldsymbol{x}_t, t) = D_\theta(\boldsymbol{x}_t, t, \boldsymbol{c}) + (w-1) (D_\theta(\boldsymbol{x}_t, t, \boldsymbol{c}) - D_\theta(\boldsymbol{x}_t, t, \boldsymbol{\varnothing}))
\end{equation*}

\textbf{CFG$^{*}$ formulation (for semantic probing):}
\begin{equation*}
    \hat{D}_\theta(\boldsymbol{x}_t, t) = D_\theta(\boldsymbol{x}_t, t, \boldsymbol{c}_{\text{deg}}) + (w-1) (D_\theta(\boldsymbol{x}_t, t, \boldsymbol{c}_{\text{deg}}) - D_\theta(\boldsymbol{x}_t, t, \boldsymbol{\varnothing}))
\end{equation*}

where $\boldsymbol{c}_{\text{deg}}$ is constructed according to Eq.~(11) in the main text with varying degradation ratios $R_{\text{deg}}$.

\subsubsection{Additional Qualitative Results.}
\label{par:appendix_cfg_star_qualitative}

\begin{figure}[h]
    \centering
    \includegraphics[width=0.9\columnwidth]{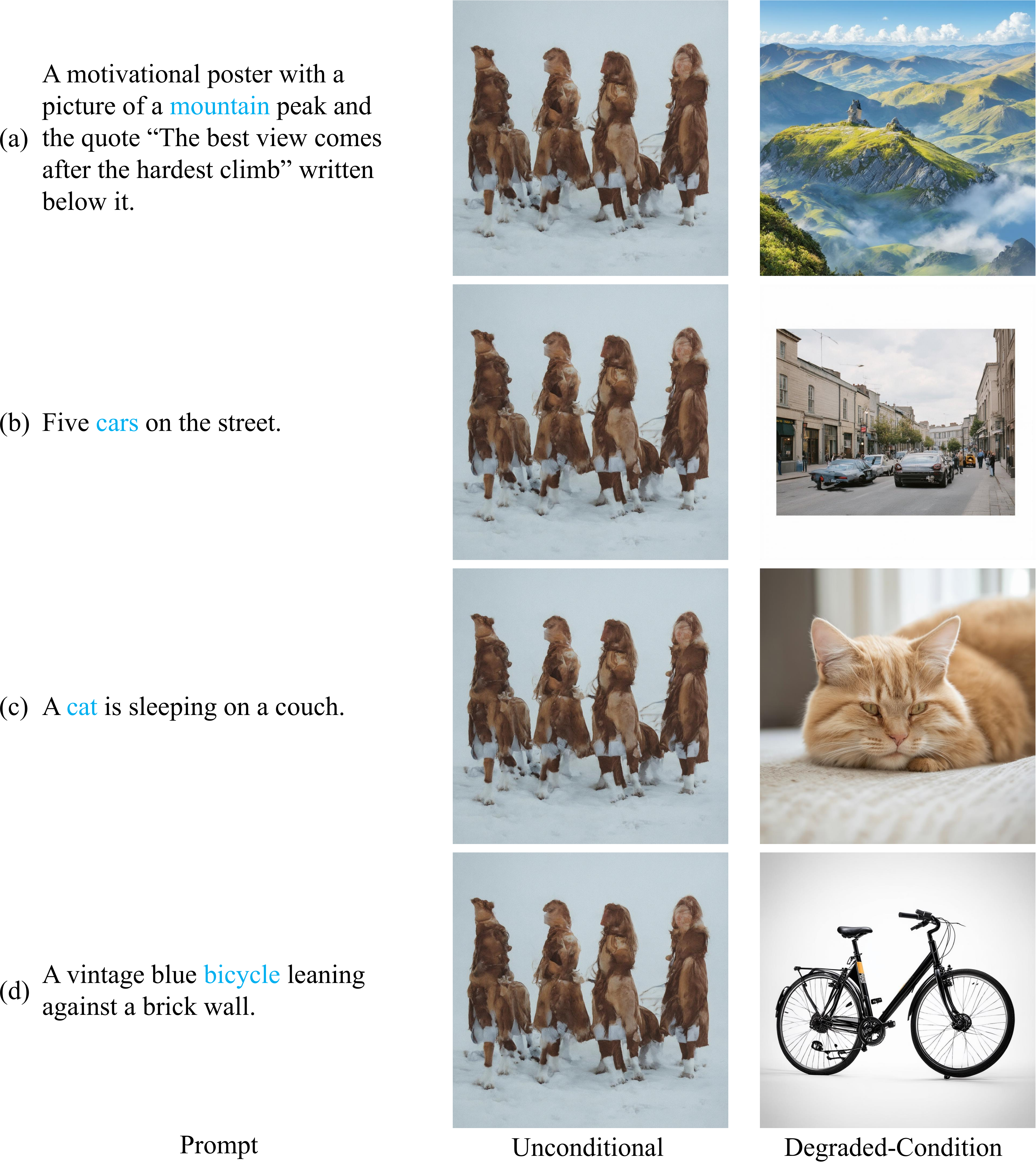} 
    \caption{Comparison between unconditional generation (left) and context-aggregating-token-only guidance (right) for four prompts. The right column uses CFG$^{*}$ with $\boldsymbol{c}_{\text{deg}}$ constructed by $r_{\text{content}}=1, r_{\text{CtxAgg}}=0$.}
    \label{fig:ctx_semantics}
\end{figure}

To provide additional visual evidence for the semantic properties of degraded conditions, Figure~\ref{fig:ctx_semantics} shows generation results guided purely by context-aggregating token information (i.e., $r_{\text{content}}=1, r_{\text{CtxAgg}}=0$). We compare unconditional generation (left) with context-aggregating-token-only guidance (right) across four diverse prompts. These results illustrate that context-aggregating tokens, after contextual encoding by the text encoder, carry global semantic information related to the original prompt.

\subsection{Full Benchmark Results on GenAI-Bench}
\label{sec:benchmark_full_results}

This section provides comprehensive quantitative results of our CDG method compared to baselines (CFG, CADS, ICG, SEG, and PAG) on the GenAI-Bench dataset, covering both advanced reasoning skills and basic skills. All metrics are reported with 2 decimal places. For each skill dimension, the best-performing method is marked in \textbf{bold}, while the second-best is \underline{underlined}.

\subsubsection{Advanced Reasoning Skills}
Table~\ref{tab:advanced_skills_full} presents detailed evaluation results for advanced reasoning skills across three state-of-the-art text-to-image diffusion models: SD3, SD3.5, and FLUX.1. These skills encompass five key reasoning capabilities: Counting, Comparison, Differentiation, Negation, and Universal reasoning. Our CDG method demonstrates consistent improvements over baseline methods across most skill dimensions.

\begin{table*}[h]
    \small
    \centering
    \begin{tabular}{clcccccc}
    \toprule
    \textbf{Model} & \textbf{Method} & \textbf{Counting} & \textbf{Comparison} & \textbf{Differentiation} & \textbf{Negation} & \textbf{Universal} & \textbf{Avg} \\
    \midrule
    \multirow{6}{*}{\textbf{SD3}} 
    & CFG & \underline{76.00} & \underline{74.08} & \underline{77.22} & 46.58 & \underline{70.74} & \underline{68.92} \\
    & CADS & 75.59 & 74.05 & 76.98 & 46.74 & 70.11 & 68.70 \\
    & ICG & 75.37 & 73.10 & 75.89 & 47.46 & 69.39 & 68.24 \\
    & SEG & 70.87 & 69.43 & 70.87 & \textbf{47.90} & 64.96 & 64.81 \\
    & PAG & 67.76 & 66.40 & 67.09 & \underline{47.84} & 64.80 & 62.78 \\
    & \textbf{CDG (Ours)} & \textbf{76.50} & \textbf{74.86} & \textbf{78.26} & 46.24 & \textbf{71.53} & \textbf{69.48} \\
    \midrule
    \multirow{6}{*}{\textbf{SD3.5}} 
    & CFG & \underline{76.45} & 73.70 & 75.10 & 46.36 & \underline{72.21} & \underline{68.76} \\
    & CADS & 76.27 & 73.54 & 75.08 & 46.63 & 71.94 & 68.69 \\
    & ICG & 76.41 & \underline{74.13} & \underline{75.63} & \underline{46.85} & 70.23 & 68.65 \\
    & SEG & 71.98 & 71.43 & 72.80 & 46.53 & 67.73 & 66.09 \\
    & PAG & 72.79 & 71.02 & 72.38 & \textbf{48.68} & 66.17 & 66.21 \\
    & \textbf{CDG (Ours)} & \textbf{77.92} & \textbf{76.06} & \textbf{78.74} & 46.65 & \textbf{73.13} & \textbf{70.50} \\
    \midrule
    \multirow{4}{*}{\textbf{FLUX.1}} 
    & CFG & \textbf{75.18} & \underline{72.97} & \underline{75.39} & \textbf{45.51} & 71.13 & \underline{68.04} \\
    & CADS & 74.84 & 72.58 & 74.87 & \underline{45.42} & 70.81 & 67.70 \\
    & ICG & 74.57 & 72.83 & 74.50 & 44.65 & \underline{71.47} & 67.60 \\
    & \textbf{CDG (Ours)} & \underline{74.98} & \textbf{73.47} & \textbf{76.17} & 44.97 & \textbf{71.55} & \textbf{68.23} \\
    \bottomrule
    \end{tabular}
    \caption{Comprehensive GenAI-Bench evaluation results for advanced reasoning skills. For each skill dimension within each model, the \textbf{best} method is shown in bold and the \underline{second-best} is underlined. CDG achieves the best performance in most dimensions, demonstrating its superior reasoning capabilities.}
    \label{tab:advanced_skills_full}
\end{table*}

\subsubsection{Basic Skills}
Table~\ref{tab:basic_skills_full} presents detailed evaluation results for basic compositional skills across SD3, SD3.5, and FLUX.1. These fundamental skills include: Action Relation (capturing action-object interactions), Attribute (object properties), Scene (environmental context), Spatial Relation (object positioning), and Part Relation (part-whole relationships). Our method shows consistent performance gains across all models.

\begin{table*}[h]
    \small
    \centering
    \begin{tabular}{clcccccc}
    \toprule
    \textbf{Model} & \textbf{Method} & \textbf{Action Relation} & \textbf{Attribute} & \textbf{Scene} & \textbf{Spatial Relation} & \textbf{Part Relation} & \textbf{Avg} \\
    \midrule
    \multirow{6}{*}{\textbf{SD3}} 
    & CFG & \underline{79.06} & \underline{77.23} & \underline{76.67} & \underline{78.66} & \underline{76.35} & \underline{77.59} \\
    & CADS & 78.84 & 77.05 & 76.51 & 78.55 & \textbf{76.36} & 77.46 \\
    & ICG & 77.95 & 76.42 & 75.77 & 78.09 & 74.96 & 76.64 \\
    & SEG & 72.57 & 71.98 & 71.28 & 73.50 & 70.99 & 72.06 \\
    & PAG & 68.58 & 68.35 & 66.95 & 69.65 & 68.75 & 68.46 \\
    & \textbf{CDG (Ours)} & \textbf{79.29} & \textbf{77.56} & \textbf{77.18} & \textbf{79.84} & 76.19 & \textbf{78.01} \\
    \midrule
    \multirow{6}{*}{\textbf{SD3.5}} 
    & CFG & \underline{79.48} & \underline{78.00} & \underline{77.45} & \underline{79.66} & \underline{76.31} & \underline{78.18} \\
    & CADS & 79.39 & 77.91 & 77.25 & 79.58 & 76.30 & 78.09 \\
    & ICG & 79.08 & 77.83 & 76.87 & 78.90 & 76.28 & 77.79 \\
    & SEG & 76.02 & 74.42 & 74.09 & 76.34 & 73.21 & 74.82 \\
    & PAG & 75.53 & 74.10 & 73.47 & 75.64 & 73.04 & 74.35 \\
    & \textbf{CDG (Ours)} & \textbf{81.18} & \textbf{79.19} & \textbf{78.18} & \textbf{80.69} & \textbf{78.08} & \textbf{79.46} \\
    \midrule
    \multirow{4}{*}{\textbf{FLUX.1}} 
    & CFG & \underline{77.33} & \underline{76.03} & \underline{76.09} & \underline{77.47} & \underline{74.78} & \underline{76.34} \\
    & CADS & 77.05 & 75.71 & 75.92 & 77.42 & 74.33 & 76.09 \\
    & ICG & 77.10 & 75.52 & 75.87 & 77.07 & 74.65 & 76.04 \\
    & \textbf{CDG (Ours)} & \textbf{77.70} & \textbf{76.27} & \textbf{76.37} & \textbf{77.56} & \textbf{74.93} & \textbf{76.57} \\
    \bottomrule
    \end{tabular}
    \caption{Comprehensive GenAI-Bench evaluation results for basic compositional skills. For each skill dimension within each model, the \textbf{best} method is shown in bold and the \underline{second-best} is underlined. CDG consistently outperforms baseline methods across different models and skill types.}
    \label{tab:basic_skills_full}
\end{table*}

\subsection{Encoder Ablation and WPR Analysis}
\label{appendix:encoder_wpr}

This section provides two supplementary ablations referenced in the main text: encoder-specific degradation analysis and WPR vs.\ random ranking comparison.

\subsubsection{Encoder-Specific Ablation}
In SD3, CLIP-L and CLIP-G are concatenated along the feature dimension (fused and inseparable), then concatenated with T5 along the sequence length dimension. Our method operates on this combined embedding, degrading CLIP and T5 portions independently. Table~\ref{tab:encoder_ablation} shows that degrading only CLIP tokens (CDG-C) yields the best FID and CLIP Score, while degrading only T5 tokens (CDG-T) provides competitive results. The full CDG (both encoders) achieves the best balance across all metrics, confirming that joint degradation is optimal.

\begin{table}[h!]
\centering
\caption{Encoder-specific ablation on SD3. -C/-T: CLIP/T5 only degradation.}
\label{tab:encoder_ablation}
\small
\setlength{\tabcolsep}{3pt}
\begin{tabular}{lccccc}
\toprule
\textbf{Method} & \textbf{FID}$\downarrow$ & \textbf{CLIP}$\uparrow$ & \textbf{Aes}$\uparrow$ & \textbf{VQA}$\uparrow$ \\
\midrule
CFG & 35.69 & 31.73 & 5.66 & 91.44 \\
CDG-C (CLIP only) & \textbf{32.68} & \textbf{32.01} & 5.65 & \underline{92.32} \\
CDG-T (T5 only) & 34.81 & 31.94 & \underline{5.67} & 92.11 \\
CDG (Both) & \underline{34.05} & \underline{32.00} & \textbf{5.70} & \textbf{92.40} \\
\bottomrule
\end{tabular}
\end{table}

\subsubsection{WPR vs.\ Random Ranking}
At the default $R_{\text{deg}}=1.0$, all content tokens are degraded regardless of ranking, making WPR unnecessary. For $R_{\text{deg}} \neq 1.0$, where only a subset of tokens is degraded, we compare WPR-based ranking against random ranking. Table~\ref{tab:wpr_ablation} shows results at $R_{\text{deg}}=0.9$ and $R_{\text{deg}}=1.1$.

\begin{table}[h!]
\small
\setlength{\tabcolsep}{2pt}
\centering
\caption{WPR vs.\ Random ranking ablation on SD3.}
\label{tab:wpr_ablation}
\begin{tabular}{c|cc|cc|cc|cc}
\toprule
\multirow{2}{*}{$R_{\text{deg}}$} & \multicolumn{2}{c|}{FID$\downarrow$} & \multicolumn{2}{c|}{CLIP$\uparrow$} & \multicolumn{2}{c|}{Aes$\uparrow$} & \multicolumn{2}{c}{VQA$\uparrow$} \\
& WPR & Rand$^*$ & WPR & Rand$^*$ & WPR & Rand$^*$ & WPR & Rand$^*$ \\
\midrule
1.1 & 33.80 & 34.17 & 31.98 & 32.02{\tiny $\pm$0.52} & 5.68 & 5.68{\tiny $\pm$0.16} & 92.27 & 92.27{\tiny $\pm$0.88} \\
0.9 & 35.08 & 33.73 & 31.72 & 31.90{\tiny $\pm$0.95} & 5.67 & 5.67{\tiny $\pm$0.28} & 91.05 & 91.65{\tiny $\pm$1.57} \\
\bottomrule
\end{tabular}
\end{table}
{\footnotesize $^*$ Means computed on COCO-5K. Std ($\pm$): variance across 20 seeds on a 400-prompt subset (FID std omitted).}

Mean performance is comparable, but WPR provides two key benefits: (1) \textbf{Stability}: random ranking introduces variance (e.g., VQA $\pm$1.57 at $R_{\text{deg}}=0.9$), while WPR ensures deterministic behavior. (2) \textbf{Theoretical grounding}: the content/context-aggregating dichotomy revealed by WPR validates $R_{\text{deg}}=1.0$ as a principled semantic boundary, providing the analytical foundation for stratified degradation.

\subsection{More Visual Results}
\label{sec:more_results}

This section presents additional visual results demonstrating the effectiveness and versatility of our CDG method. We first showcase CDG's compatibility with orthogonal techniques, followed by comprehensive qualitative comparisons across different models.

\subsubsection{Combination with Orthogonal Methods.}
\begin{figure}[h]
    \centering
    \includegraphics[width=0.7\columnwidth]{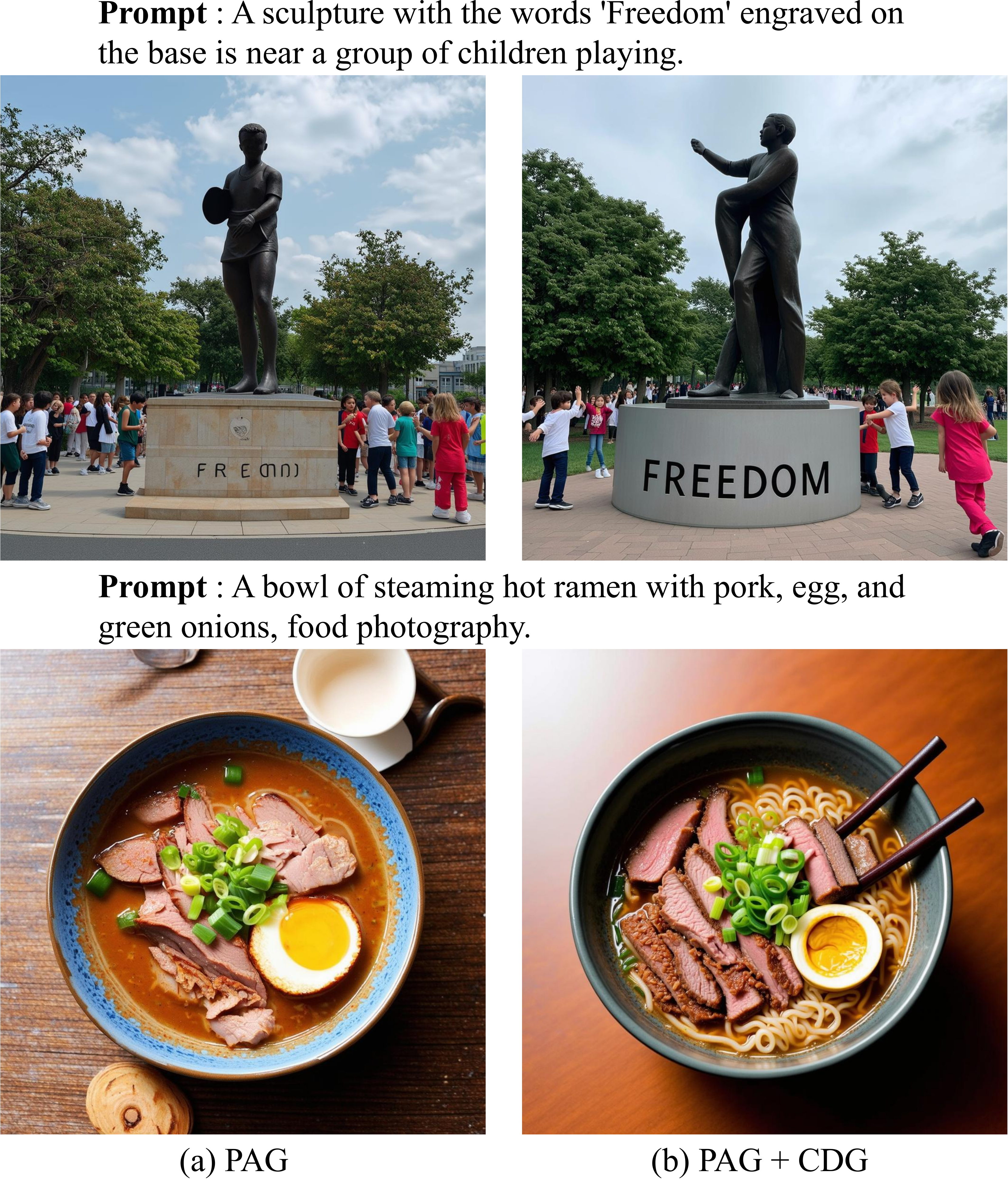}
    \caption{Synergy with PAG: CDG provides the core semantic foundation, while PAG refines local details.}
    \label{fig:cdg_pag}
\end{figure}

The synergy of CDG with orthogonal methods like PAG is shown in Figure~\ref{fig:cdg_pag}. In the sculpture example, PAG alone fails to correctly render the word ``Freedom" on the base, producing a garbled text ``FRE(nn)". With CDG, the semantic constraint ensures accurate text generation, displaying the complete and correct ``FREEDOM" inscription. Similarly, for the food photography prompt, PAG alone struggles to capture the style specification, producing an image with distracting background clutter. The addition of CDG provides the necessary semantic precision, removing this clutter and results in a cleaner, more professional composition.
\subsubsection{Application to Text-Guided Image Editing.}
\begin{figure}[h]
    \centering
    \includegraphics[width=0.9\columnwidth]{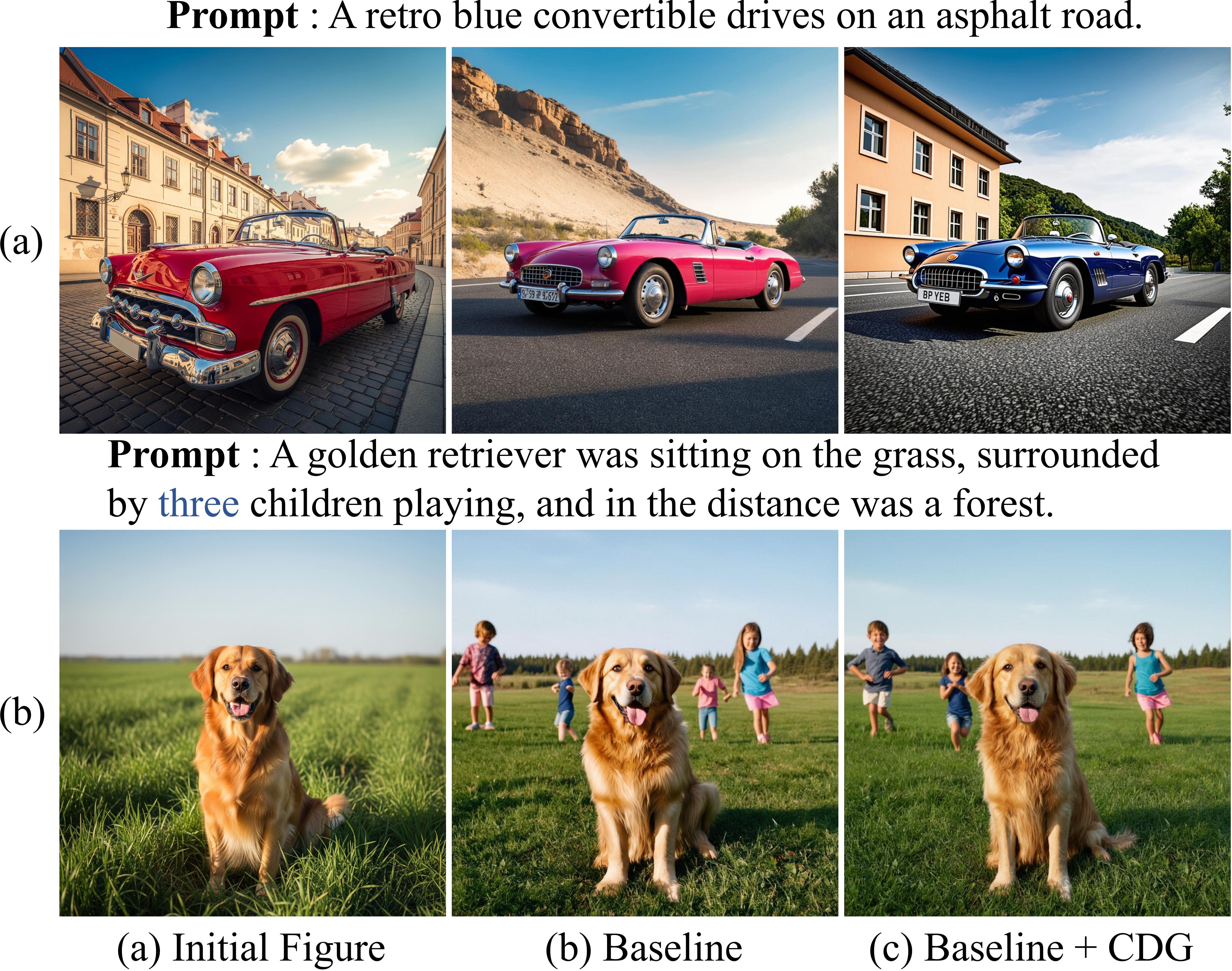}
    \caption{CDG for Text-Guided Editing: Correcting semantic errors where the baseline fails.}
    \label{fig:cdg_i2i}
\end{figure}

CDG's semantic precision is particularly beneficial for text-guided image editing tasks. 
As shown in Figure~\ref{fig:cdg_i2i}, in the convertible car example, the baseline struggles with the color attribute, generating a red or pink car instead of the specified ``blue". CDG corrects this semantic error, producing the accurate retro blue convertible as described in the prompt. Similarly, in the golden retriever scene, while the baseline method preserves the overall image layout, it fails to interpret a key numerical constraint in the prompt (``three children"). 
In contrast, CDG successfully enforces this semantic detail, rendering the correct number of subjects without disrupting the composition.

\subsubsection{Application to Controllable Generation.}
\begin{figure}[h]
    \centering
    \includegraphics[width=0.9\columnwidth]{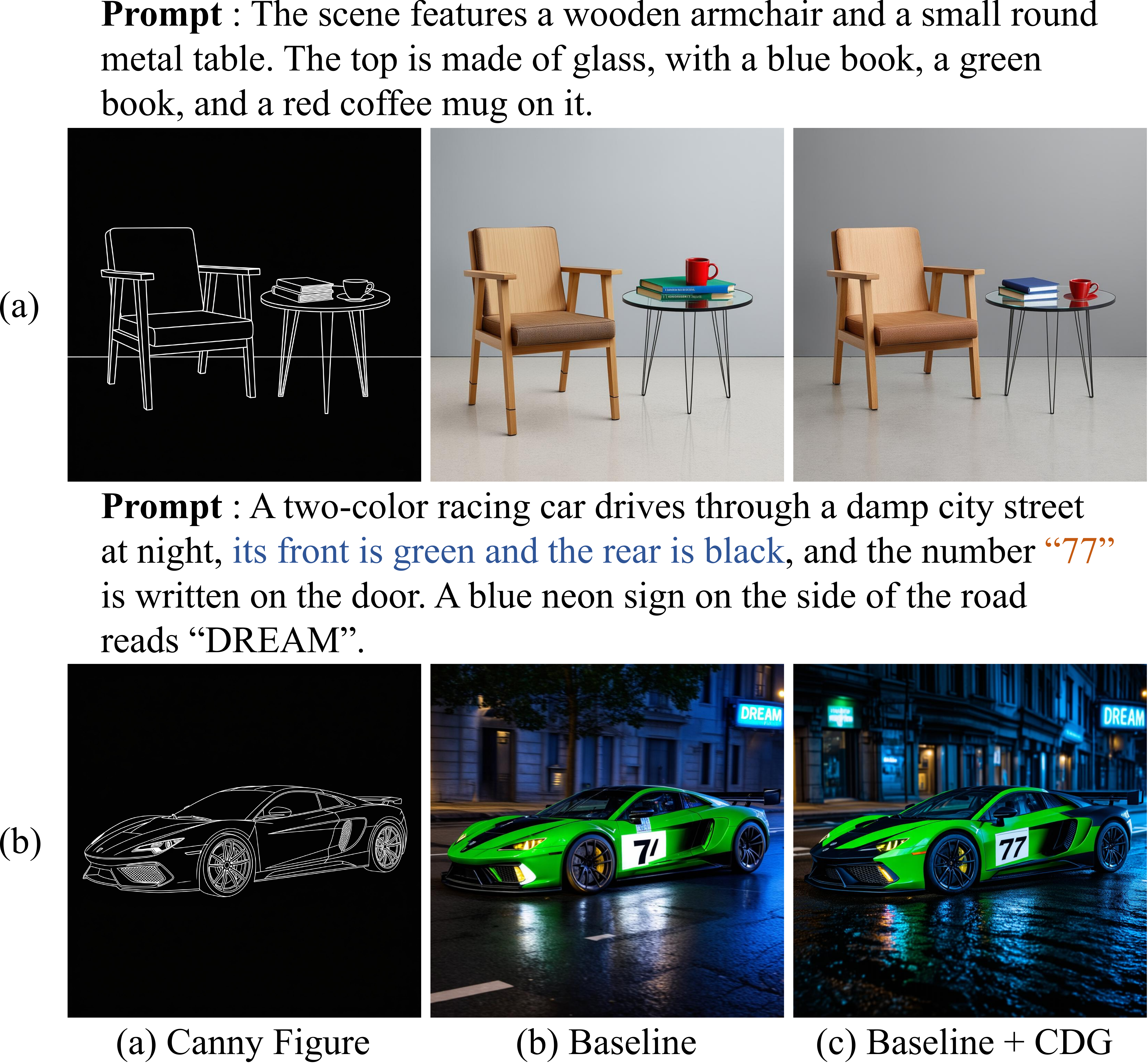}
    \caption{CDG for Controllable Generation: Providing semantic precision to render complex details under structural constraints.}
    \label{fig:cdg_controlnet}
\end{figure}

As shown in Figure~\ref{fig:cdg_controlnet}, in the furniture scene, ControlNet alone fails to follow both the structural layout and color attributes. The baseline incorrectly places the coffee mug on top of the books, whereas the canny figure shows it should be positioned to the right of the books. Additionally, the colors are wrong (``blue book, green book, red coffee mug" are not rendered correctly). With CDG, the model correctly positions all objects according to the structural constraint and renders them with their specified colors. Similarly, in the racing car example, while ControlNet successfully enforces structural constraints (from a canny figure), it struggles with complex combinatorial semantics (e.g., ``green front, black rear" and ``77"). 
By providing a more precise semantic signal, CDG enables the model to faithfully render these details, demonstrating a powerful synergy between structural and semantic control.

\subsubsection{More Visual Results on SD3, SD3.5, and FLUX.1}

\begin{figure*}[h]
    \centering
    \includegraphics[width=0.87\textwidth]{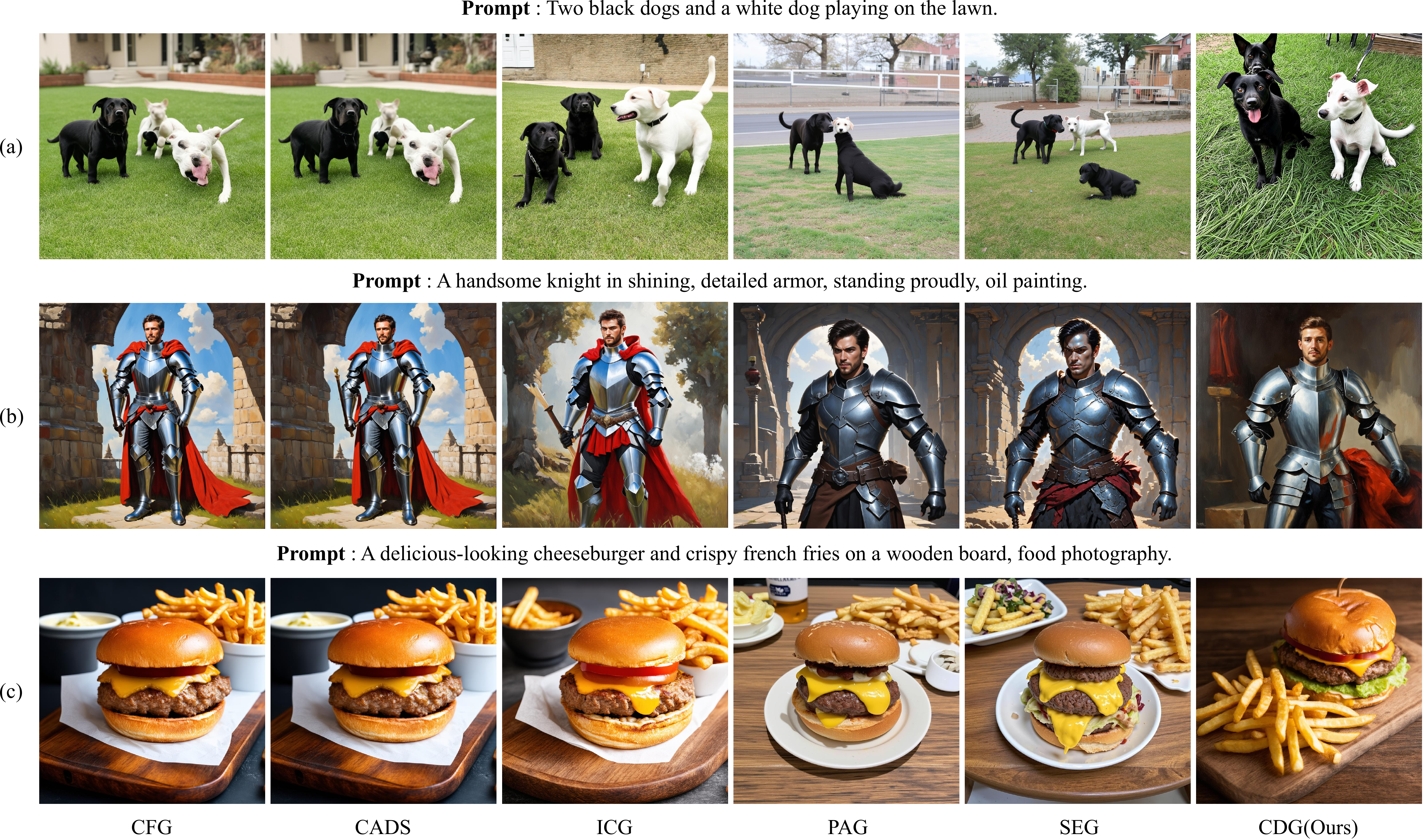}
    \caption{More visual results on the SD 3 model.}
    \label{fig:more_results_sd3}
\end{figure*}

\begin{figure*}[h!]
    \centering
    \includegraphics[width=0.87\textwidth]{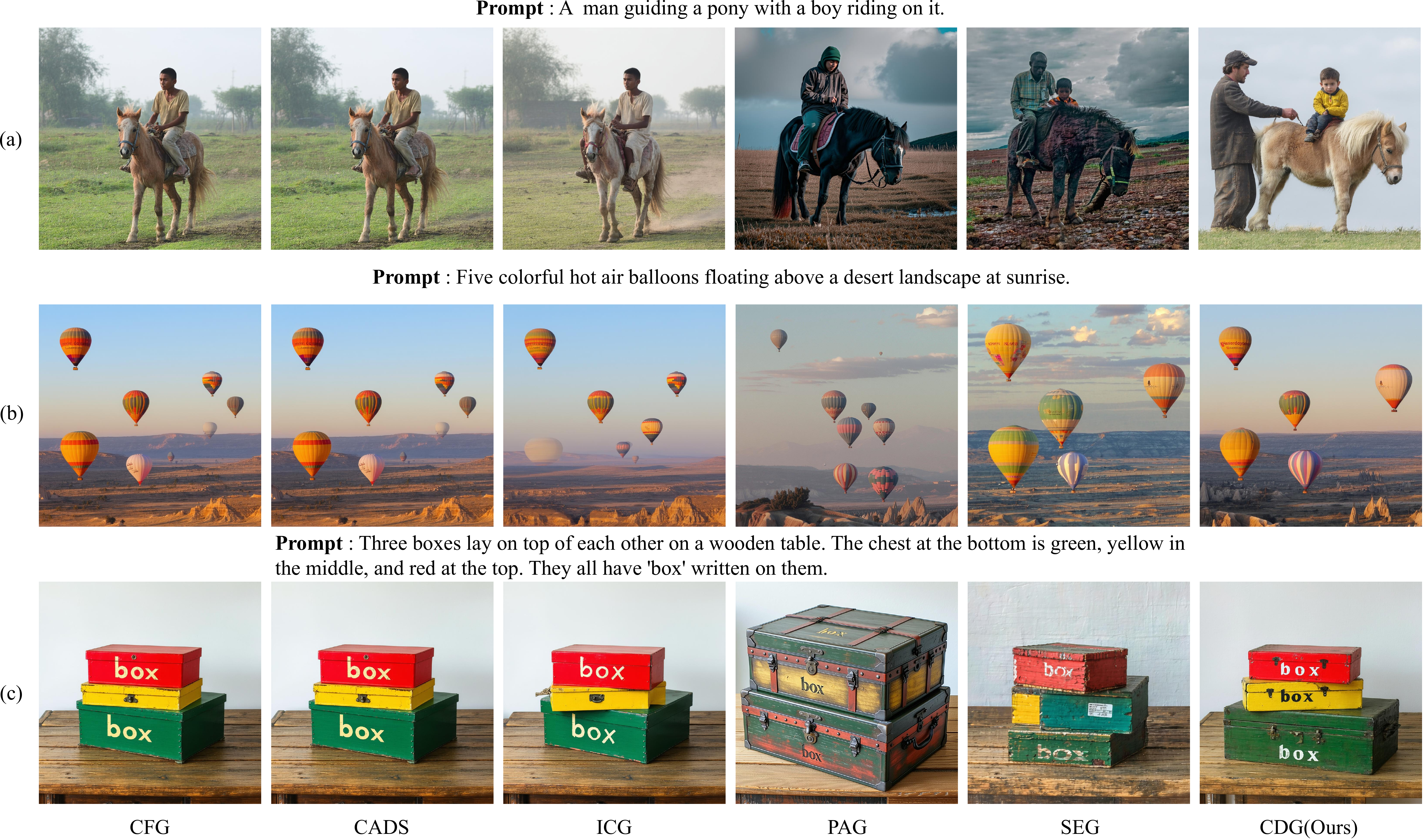}
    \caption{More visual results on the SD 3.5 model.}
    \label{fig:more_results_sd3.5}
\end{figure*}

\begin{figure*}[h!]
\centering
\includegraphics[width=0.87\textwidth]{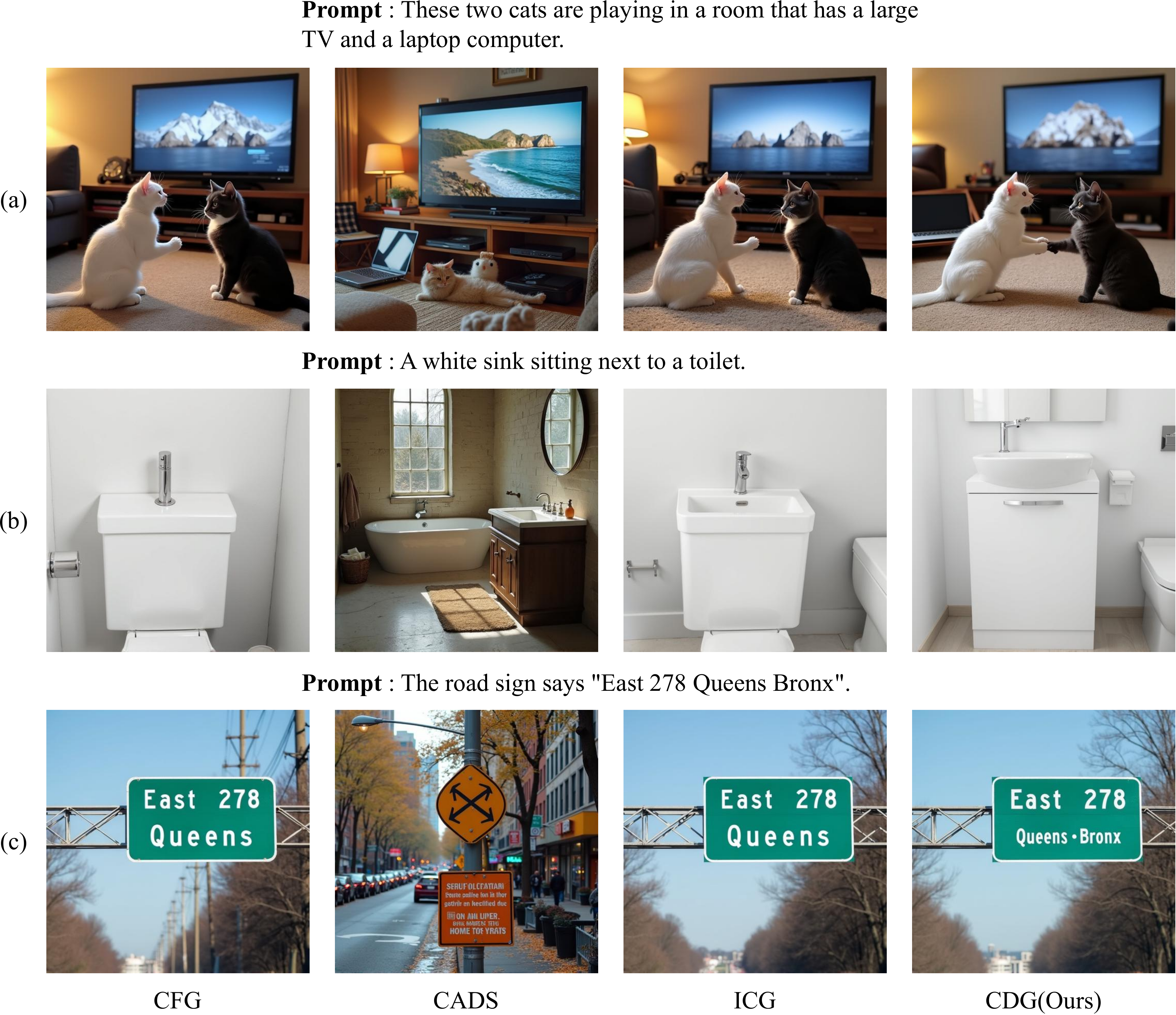}
\caption{More visual results on the Flux model.}
\label{fig:more_results_flux}
\end{figure*}

\end{document}